\documentclass[10pt, letter, onecolumn]{arxiv}

\usepackage{kantlipsum, lipsum}
\usepackage{dm-colors}
\usepackage{amsmath}
\usepackage{pstricks, pst-node}
\usepackage{verbatim}
\usepackage{multirow}
\usepackage{scalerel}
\usepackage{booktabs}
\usepackage{enumitem}
\usepackage{xspace}
\usepackage{bm}
\usepackage{bbm}
\usepackage{mathtools}
\usepackage{soul}
\usepackage{epsfig}
\usepackage{graphicx}
\usepackage{amssymb}
\usepackage{colortbl}
\usepackage{csquotes}
\usepackage{setspace}
\usepackage{colortbl}
\usepackage{bbding}
\usepackage{threeparttable}
\usepackage{tabularx,ragged2e}
\usepackage{placeins}
\usepackage{bbding}
\definecolor{darkpastelgreen}{rgb}{0.13, 0.55, 0.13}
\definecolor{darkpastelred}{rgb}{0.55, 0.13, 0.13}
\usepackage[hang,flushmargin]{footmisc}

\usepackage{nameref}
\usepackage{varioref}
\usepackage{amssymb}% 
\usepackage{lineno}
\usepackage{pifont}% http://ctan.org/pkg/pifont
\newcommand{\xmark}{\ding{55}}%

\usepackage[pagebackref=false,breaklinks=false,%
            colorlinks=true,bookmarks=true,citecolor=ourdarkblue,%
            urlcolor=ourdarkblue,linkcolor=ourdarkblue]{hyperref}
\usepackage[noabbrev,capitalize]{cleveref}
\usepackage{etoc}
\graphicspath{{figures/}}
\definecolor{midgreen}{rgb}{0.0, 0.75, 0.0}

\newcommand{\change}[1]{{\textcolor{black}{#1}}}

\title{\Large{Large-scale Long-tailed Disease Diagnosis on Radiology Images}}

\author[$\ast$,1,2]{\hspace{4pt}Qiaoyu Zheng} %#1
\author[$\ast$,1,2]{Weike Zhao} %#1
\author[$\ast$,1,2]{Chaoyi Wu} %#1
\author[1,2]{Xiaoman Zhang} %#1
\author[1,3]{Lisong Dai}
\author[4,5]{Hengyu Guan}
\author[1,3]{\\ \vspace{0.1cm} \hspace{2pt} Yuehua Li}
\author[1,2]{Ya Zhang}
\author[1,2,$\dag$]{Yanfeng Wang} 
\author[1,2,$\dag$]{Weidi Xie}

\affil[1]{\normalsize Shanghai Jiao Tong University \hspace{1cm}}

\affil[2]{\normalsize Shanghai AI Laboratory \authorcr \vspace{2pt}}

\affil[3]{\normalsize Shanghai Sixth People’s Hospital, Affiliated to Shanghai Jiao Tong University \authorcr \vspace{2pt}}

\affil[4]{\normalsize Department of Reproductive Medicine, Ren Ji Hospital, Shanghai Jiao Tong University School of Medicine \authorcr \vspace{2pt}}

\affil[5]{\normalsize Shanghai Key Laboratory for Assisted Reproduction and Reproductive Genetics  \authorcr \vspace{2pt}
\url{https://qiaoyu-zheng.github.io/RP3D-Diag}
}

\renewcommand{\correspondingauthor}[1]{$\ast$~Equal contributions. \\ $\dag$~Corresponding author. Email addresses: \{three-world, zwk0629, wtzxxxwcy02, weidi\}@sjtu.edu.cn}

\begin{document}
%\begin{refsection}
% \linenumbers 

\begin{abstract}
Developing a generalist radiology diagnosis system can greatly enhance clinical diagnostics.  
In this paper, we introduce \textbf{RadDiag}, a foundational model supporting 2D and 3D inputs across various modalities and anatomies, using a transformer-based fusion module for comprehensive disease diagnosis. 
Due to patient privacy concerns and the lack of large-scale radiology diagnosis datasets, we utilize high-quality, clinician-reviewed radiological images available online with diagnosis labels. 
% Considering patient privacy, no existing dataset enables research on large-scale radiology disease diagnosis, we exploit the online highly specialized, high-quality labeled radiological images reviewed by experienced clinicians. 
Our dataset, \textbf{RP3D-DiagDS}, contains \change{40,936 cases with 195,010 scans} covering 5,568 disorders (930 unique ICD-10-CM codes).
% Specifically, this dataset, namely \textbf{RP3D-DiagDS}, includes 39,026 cases with 192,675 scans, 
% spanning over 5,568 disorders~(associated with 930 unique ICD-10-CM codes).
Experimentally, our \textbf{RadDiag} achieves \change{95.14\% AUC} on internal evaluation with the knowledge-enhancement strategy. 
Additionally, \textbf{RadDiag} can be \change{zero-shot applied or fine-tuned} to external diagnosis datasets sourced from various hospitals, demonstrating state-of-the-art results. In conclusion, we show that publicly shared medical data on the Internet is a tremendous and valuable resource that can potentially support building a generalist AI for healthcare.

\end{abstract}

\maketitle

% ------------ SECTIONS ---------------------

% ========== Edit your name here
\section{INTRODUCTION}

In the landscape of clinical medicine, the emergence of radiology techniques, for example, X-ray, CT, and MRI, has truly revolutionized the medical field, which enables to diagnose by reading radiological images. Recent literature has demonstrated significant potential for developing AI models for disease diagnosis in the medical field~\cite{zhang2023knowledge,tiu2022expert}. 
However, a notable limitation of the existing models is their specialization, as these `specialist' models often focus on a narrow range of disease categories, and thus usually fail to accommodate the breadth and complexity of clinical scenarios. Conversely, an ideal radiology diagnosis system is expected to comprehensively analyze the combination of arbitrary historical scans across any anatomical regions and imaging modalities for a wide range of diseases. 

In this paper, we aim to build a foundational radiology diagnostic model, that supports information fusion from an arbitrary number of 2D and 3D inputs across different modalities and anatomic sites. 
Generally speaking, there are three major challenges. \textit{First}, lack of datasets for training. Despite the large volume of radiology data generated per day across the world, 
publicly available ones are extremely limited due to privacy issues.
Existing open-source datasets are normally collected from a certain modality or anatomy with very limited diseases, thus significantly hindering the development of powerful disease diagnosis models. \textit{Second}, lack of unified architecture formulation. Models for diagnosing radiological diseases are often tailored to particular datasets, which are optimized for 
a narrow spectrum of diseases linked with particular imaging modalities~\cite{dai2021transmed,irvin2019chexpert}.
In contrast, developing a foundation model for disease diagnosis requires the architecture to be capable of fusing information from an arbitrary number of images of various modalities, for example, X-ray, CT, and MRI.  \textit{Third}, lack of benchmark to monitor progress. Benchmarking the models' performance on disease diagnosis relies on datasets that well simulate real-world scenarios, for instance, containing many diseases, long-tail distribution, {\em etc}. Existing benchmarks typically focus on a very limited number of diseases, with single modality inputs~\cite{bien2018deep,majkowska2020chest}. 
To comprehensively measure the progress in disease diagnosis, 
a high-quality benchmark has yet to be established.

% \vspace{-0.25cm}
% \begin{itemize}
% \setlength\itemsep{0.1cm}
% \item {\bf Lack of datasets for training:} despite the large volume of radiology data generated per day across the world, 
% publicly available ones are extremely limited due to privacy issues.
% Existing open-source datasets~\cite{jack2008alzheimer} are normally collected from a certain modality or anatomy with very limited diseases, thus significantly hindering the development of powerful disease diagnosis models.

% \item {\bf Lack of unified architecture formulation:}
% models for diagnosing radiological diseases are often tailored to particular datasets, which are optimized for 
% a narrow spectrum of diseases linked with particular imaging modalities~\cite{dai2021transmed,cai2019breast}. 
% In contrast, developing a foundation model for disease diagnosis requires the architecture to be capable of fusing information from an arbitrary number of images of various modalities, for example, X-ray, CT, and MRI. 

% \item {\bf Lack of benchmark to monitor progress:} 
% benchmarking the models' performance on disease diagnosis relies on datasets that well simulate real-world scenarios, for instance, containing a large number of diseases, long-tail distribution, {\em etc}. Existing benchmarks typically focus on a very limited number of diseases, with single modality inputs~\cite{bien2018deep,majkowska2020chest}. 
% To comprehensively measure the progress in disease diagnosis, 
% a high-quality benchmark is yet to be established.
% \end{itemize}

\begin{figure}[!t]
    \centering
    \includegraphics[width=1\linewidth]{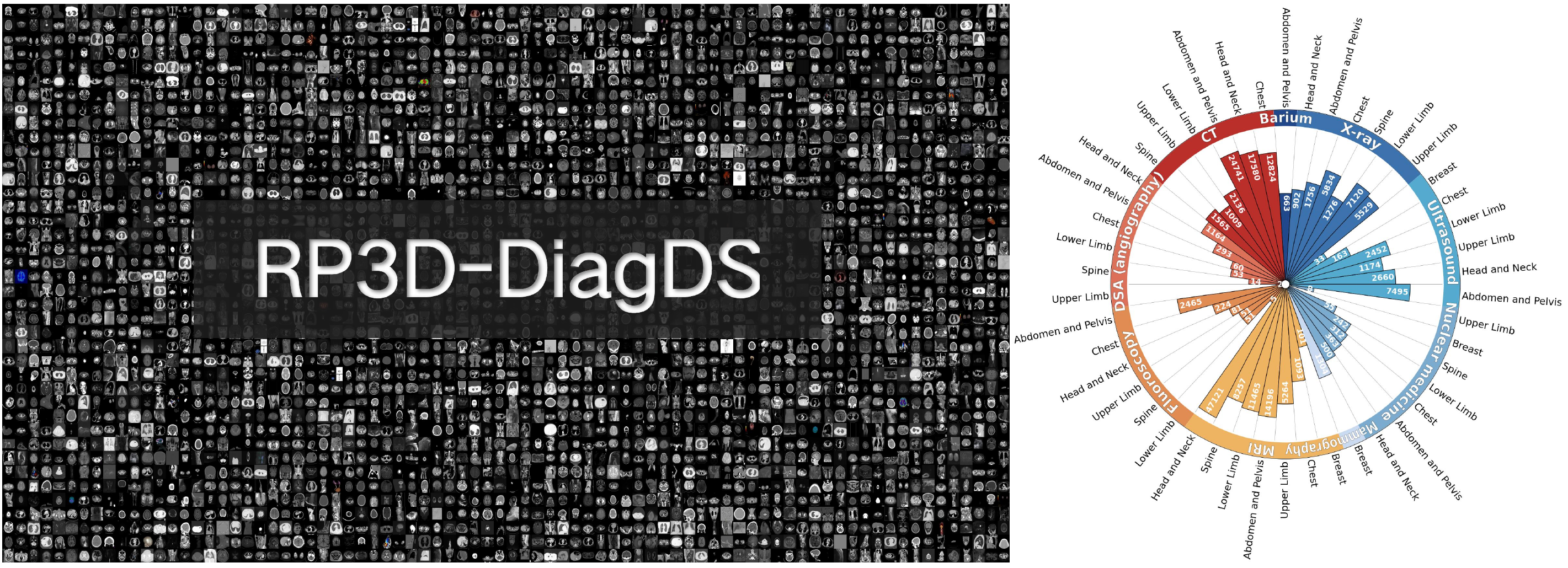}
    \caption{\textbf{Overview of RP3D-DiagDS}. There are \textbf{39,026 cases (192,675 scans)} across 7 human anatomy regions and 9 diverse modalities covering \textbf{930 ICD-10-CM codes}.}
    \label{fig:teaser}
    \vspace{-15pt}
\end{figure}

To overcome the above-mentioned challenges, 
we consider to exploit the widely available data on Internet, especially those focusing on high-quality radiology samples from professional institutions and educational websites, that have undergone meticulous de-identification, and reviewed by a panel of experienced clinicians. Specifically, we develop a flexible data collection pipeline, with each sample associated with an International Classification of Diseases code, {\em i.e.}, ICD-10-CM~\cite{ICD10}, indicating the diagnostic category.
The dataset presents a long-tailed distribution across disease categories. The case numbers range from 1 to 964 with some rare diseases. It includes multi-modal scans per case, with modalities varying from 1 to 5 and image counts between 1 and 30. 
As a result, we collect a multi-scan medical disease classification dataset, with \change{\textbf{40,936 cases (195,010 scans)}} across \textbf{7} anatomy regions and \textbf{9} modalities, covering \textbf{930 ICD-10-CM codes and 5568 disorders}, including a broad spectrum of medical conditions.

Architecturally, we conduct preliminary explorations on the design that supports both 2D and 3D input from various modalities, 
together with a transformer-based fusion module for comprehensive diagnosis. \change{Specifically, for visual encoding, we explore three variants of backbones, namely, ResNet-based, ViT-based and ResNet-ViT-mixing to perform 2D or 3D unified encoding.} 
Then, we fuse the multi-scan information with a transformer-based fusion module, treating each image embedding as an input token. 
At training time, we adopt knowledge-enhanced training strategy~\cite{Wu2023KDiagKD,zhang2023knowledge}, 
where we first leverage the rich domain knowledge to pre-train a knowledge encoder with natural language, and use it to guide the visual representation learning for disease diagnosis. 

In experiments, we evaluate our model on a benchmark, constructed by images from online resources, featuring large-scale, multi-modal, multi-scan, long-tailed disease diagnosis. Additionally, we demonstrate that our proposed model can function as a general-purpose image encoder to capture a better image representation for radiology, making it possible to train and classify new sets via data-efficient \change{fine-tuning or zero-shot transferring}, leading to improved performance across diverse disease diagnosis tasks on \change{numerous} datasets, regardless of their image dimensions, modalities and anatomies. This paper demonstrates that publicly shared medical information is a tremendous resource that can be harnessed to advance the development of general AI for healthcare.

\section{RESULTS}
In this section, we start by briefly introducing our considered problem formulation, followed with showing the characteristics of our dataset for case-level multi-modal, multi-scan, long-tailed disorder/disease diagnosis. \change{Then, we leverage a test split of the dataset to carry out a series of studies on diagnostic performance and establish a case-level diagnosis benchmark.} 
Lastly, we train and assess the developed model accordingly, along with its transferring ability to various external datasets.

\subsection*{Problem Formulation}
\label{problem_formulation}

%\weidi{I think we may need to bring the problem formulation here, otherwise, the readers may not be able to understand the results.}

%\weidi{we need some sentences to describe the motivation, for example, we are interested in disease diagnosis, aim to design architecture that supports to fuse information from arbitrary number of images, taken by different modalities, for example, MRI....}\qiaoyu{fixed}

In this paper, our objective is to develop a radiology diagnostic model capable of integrating information from various images within a single \textbf{case}. We define a \textbf{case} as encompassing a collection of both 2D and 3D radiological images from different modalities and anatomical sites of a patient, who may be diagnosed with multiple conditions.

More specifically, we represent a case and its associated disease labels as $\mathcal{X} = \{x_1, \dots, x_S, y_1, \dots, y_c\}$, where each $x_i$ signifies a specific scan~(either 2D or 3D) from a radiological examination such as MRI, CT, or X-ray.
Here, $S$ is the number of scans for the case, and $c$ is the total number of diagnostic labels. It is noteworthy that a patient's case might include multiple scans across the same or different modalities.

To illustrate this concept, consider a case where a patient undergoes a physical examination, including a chest CT and a chest X-ray as shown in Figure~\ref{fig:method}. In this instance, $S=2$ reflects the two scans involved in this case. If the patient is diagnosed with nodule and pneumothorax, the labels for these diseases are set to 1, reflecting their presence in the patient's diagnosis.

Our goal is to train a model that can solve the above multi-class, multi-scan diagnosis problem:
\begin{align}
\mathcal{Y} = \Phi(\mathcal{X}) = \Phi_{\text{cls}}(\Phi_{\text{fuse}}(\Phi_{\text{visual}}(x_1), \Phi_{\text{visual}}(x_2), \cdots, \Phi_{\text{visual}}(x_S))) \in \mathcal{R}^c,
\end{align}
where the model is composed of a visual encoder $\Phi_{\text{visual}}$, a fusion module $\Phi_{\text{fuse}}$ and a classifier $\Phi_{\text{cls}}$.
% For instances where the radiology images are below or exceed the standard size, the model employs cropping or padding techniques to normalize the inputs.\

\subsection*{\change{RP3D-DiagDS Statistics}}
Here, we provide a detailed statistical analysis of our collected dataset, namely RP3D-DiagDS, from three aspects: modality coverage, 
anatomy coverage, and disease coverage. As shown in Supplementary Section E, compared with existing datasets, our dataset has great advantages in terms of anatomy richness, modal richness, and the number of images and categories. Furthermore, we split the dataset based on the number of cases on different levels.

\begin{figure}[!ht]
    \centering
    \includegraphics[width=1\linewidth]{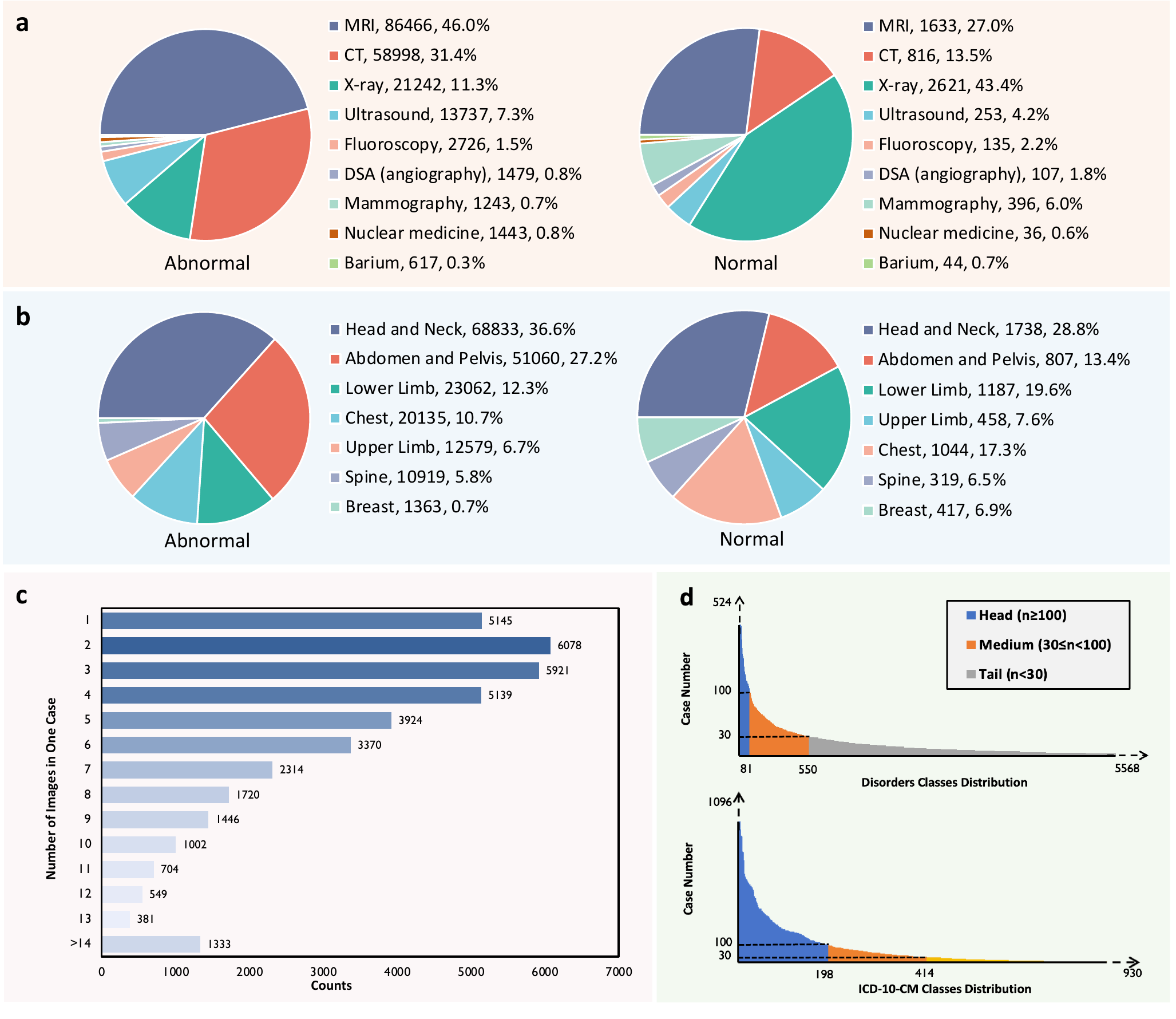}
    \caption{\textbf{The data distribution analysis on RP3D-DiagDS.}  \change{\textbf{a} The distribution of imaging modalities of abnormal (left) and normal (right) cases in RP3D-DiagDS. Each label is annotated with the class name, number of cases, and the corresponding proportion.
    \textbf{b} The distribution of imaging  anatomies of abnormal (left) and normal (right) cases in RP3D-DiagDS.} 
    \textbf{c} Case distribution on image numbers. 
    In the bar plot, We show the distribution for the number of images in one case. In RP3D-DiagDS, each case may include multiple images from patient history scans, different modalities, and different angles or conditions. \textbf{d} Case distribution on classes. 
    We demonstrate the long-tailed distributions for disorder and ICD-10-CM  classes. We also categorize these classes into three categories: ``head class'', ``body class'' and ``tail class'' based on the number of cases. Notably, to better show the main part of the case distributions, we clip the axes, indicated by the dotted axes lines.}
    \label{fig:multi-img}
\end{figure}

% \subsubsection{Data Statistics}
\noindent \textbf{Analysis on Modality Coverage.}
RP3D-DiagDS comprises images from 9 modalities, namely, computed tomography (CT), magnetic resonance imaging (MRI), X-ray, Ultrasound, Fluoroscopy, Nuclear medicine, Mammography, DSA (angiography), and Barium Enema. Each case may include images from multiple modalities, to ensure the precise and comprehensive diagnosis of disorders. Overall, approximately 19.4\% of the cases comprise images from two modalities, while around 2.9\% involve images from three to five modalities. The remaining cases are associated with image scans from a single modality. The distribution of modalities among all abnormal and normal samples are illustrated in Figure~\ref{fig:multi-img}a. 

% \begin{figure}[!t]
% \centering
% \begin{subfigure}{0.48\textwidth}
%     \includegraphics[width=\textwidth]{figure/dataset/Abnormal_Samples_Distribution_in_Modality.pdf}
%     \caption{Abnormal Cases Distribution in Modality}
%     \label{fig:111}
% \end{subfigure}
% \hfill
% \begin{subfigure}{0.48\textwidth}
%     \includegraphics[width=\textwidth]{figure/dataset/Normal_Samples_Distribution_in_Modality.pdf}
%     \caption{Normal Cases Distribution in Modality}
%     \label{fig:222}
% \end{subfigure}

% \begin{subfigure}{0.48\textwidth}
%     \includegraphics[width=\textwidth]{figure/dataset/Abnormal_Samples_Distribution_in_Anatomy.pdf}
%     \caption{Abnormal Cases Distribution in Anatomy}
%     \label{fig:333}
% \end{subfigure}
% \hfill
% \begin{subfigure}{0.48\textwidth}
%     \includegraphics[width=\textwidth]{figure/dataset/Normal_Samples_Distribution_in_Anatomy.pdf}
%     \caption{Normal Cases Distribution in Anatomy}
%     \label{fig:444}
% \end{subfigure}

% \vspace{1pt}
%     \caption{The distribution of imaging modalities of abnormal (a) and normal (b) cases in RP3D-DiagDS and anatomies of abnormal (c) and normal (d) cases in RP3D-DiagDS. Each label is annotated with the class name, number of cases, and the corresponding proportion.}
%     \label{fig:Modality_1}
% \end{figure}

\vspace{3pt}
\noindent \textbf{Analysis on Anatomy Coverage.}
RP3D-DiagDS comprises images from various anatomical regions, including the head and neck, spine, chest, breast, abdomen and pelvis, upper limb, and lower limb, providing comprehensive coverage of the entire human body. The statistics are shown in Figure~\ref{fig:multi-img}b.

\vspace{3pt}
\noindent \textbf{Analysis on Disease Coverage.}
For both disorder and disease classification, each case can correspond to multiple disorders, resulting in RP3D-DiagDS a multi-label classification dataset. As shown in Figure~\ref{fig:multi-img}d, the distributions exhibit extremely long-tailed patterns, rendering such 2D \& 3D image classification problems as a long-tailed multi-label classification task. Totally there are 5568 disorders mapped into 930 ICD-10-CM classes. We define the ``head class'' category with case counts greater than 100, the ``body class'' category with case counts between 30 and 100, and the ``tail class'' category with case counts less than 30.
% \weidi{can we include some numbers here, for example, how many disorders, that are mapped into how many ICDs, etc.}

\subsection*{\change{Ablation Studies}}
\change{We first carry out a series of ablation studies to explore the optimal model architecture and parameter configurations, including the fusion strategy, visual encoder architecture, the depth of 3D scans, and augmentation. \textbf{Note that}, considering the original \textbf{RP3D-DiagDS} is too large to carry out extensive empirical experiments, we construct a subset by only keeping the top 200 classes with the most cases, accordingly, tracking the performance on these 200 classes for ablation studies. }

\change{As simple experiment baselines, 
we use images with spatial resolution of 512x512, 16 as the 3D scan depth, and ResNet-18+ResNet-34 as the visual backbone, without knowledge enhancement or augmentation strategy, adopting the basic max pooling fusion strategy introduced in Supplementary Section B. While conducting an ablation study on certain component, we keep other settings unchanged. More results can be found in Supplementary Section C. Herein, we mainly describe the critical results on comparing image encoder architectures.}

\begin{table}[htb]
  \centering
  \small
  \setlength{\tabcolsep}{4pt}
      \caption{\textbf{Ablation study on various settings of visual encoder architectures.} ``Normalisation'' denotes the separated visual encoder part to perform 2D and 3D normalization and ``Shared Enc.'' denotes the shared encoder part for both 2D and 3D scans.}
  \vspace{3pt}
\begin{tabular}{c|cc|ccccccc}
\toprule
\multirow{2}{*}{Visual Encoder} &
  \multicolumn{2}{c|}{Architecture} &
  \multirow{2}{*}{AUC} &
  \multirow{2}{*}{AP} &
  \multirow{2}{*}{F1} &
  \multirow{2}{*}{MCC} &
  \multirow{2}{*}{R@F0.01} &
  \multirow{2}{*}{R@F0.05} &
  \multirow{2}{*}{R@D0.1} \\ \cmidrule{2-3}
                                 & \multicolumn{1}{c|}{Normalisation}   & Shared Enc.   &       &       &       &       &       &                &       \\ \midrule
\multirow{4}{*}{ViT}             & \multicolumn{1}{c|}{2-layer MLP} & 6-layer ViT  & 79.98 & 6.01  & 13.57 & 14.69 & 17.78 & 33.56          & 47.21 \\
                                 & \multicolumn{1}{c|}{4-layer MLP} & 6-layer ViT  & 80.57 & 6.13  & 13.49 & 14.78 & 17.68 & 34.01          & 47.44 \\
                                 & \multicolumn{1}{c|}{2-layer MLP} & 12-layer ViT & 81.69 & 6.40  & 14.71 & 15.30 & 18.11 & 34.73          & 48.84 \\
                                 & \multicolumn{1}{c|}{4-layer MLP} & 12-layer ViT &     82.03  &   6.67    &  14.94     &  15.66     &  18.20     &   34.99             &  49.52     \\ \midrule
\multirow{4}{*}{{ResNet}} & \multicolumn{1}{c|}{ResNet-18}   & ResNet-18    & 86.91 & 11.00 & 16.77 & 18.63 & 20.42 & 41.87          & 59.38 \\
                                 & \multicolumn{1}{c|}{ResNet-34}   & ResNet-18    & 86.99 & 11.15 & 17.14 & 19.21 & 20.82 & 44.67 & 61.13 \\
 &
  \multicolumn{1}{c|}{ResNet-18} &
  ResNet-34 &
  87.06 &
  11.27 &
  17.36 &
  19.23 &
  21.48 &
  44.38 &
  61.54 \\
& \multicolumn{1}{c|}{ResNet-34}   & ResNet-34    &  87.10    &  11.31  &  17.66   &   19.41    &  21.33     &     44.19           &   62.25    \\ \midrule
\multirow{4}{*}{\change{\textbf{ResNet-ViT}}} & \multicolumn{1}{c|}{ResNet-34}  & 6-layer ViT  & 88.74 & 11.52 & 17.86 & 20.05 & 21.92 & 44.63  & 63.09 \\
                                 &  \multicolumn{1}{c|}{ResNet-50}  & 6-layer ViT  &89.53 & \textbf{11.75} & 19.59 & 20.61 & \textbf{23.18} & 51.34 & 67.39  \\
 & \multicolumn{1}{c|}{ResNet-34} & 12-layer ViT &88.93 &11.49 &18.07 &20.09 &22.38 &45.23 &65.04 \\
& \multicolumn{1}{c|}{ResNet-50}  & 12-layer ViT  &   \textbf{89.56}  &11.73     & \textbf{19.73}   &   \textbf{20.11} & 22.58      & \textbf{51.64 }           &   \textbf{67.92 }   \\ 
\bottomrule
\end{tabular}
\label{tab:Visual Structure Comparison Main}
  % \vspace{-2mm}
\end{table}

\change{Specifically, we investigate three backbone architectures variants for realizing unified diagnosis with arbitrary 2D and 3D scans, \emph{i.e.}, ResNet-based, ViT-based and ResNet-ViT-mixing. More details on architecture design can be found in Section~\ref{architecture}.  
As shown in Table~\ref{tab:Visual Structure Comparison Main},
we make three observations, (i)~The performance of ResNet-ViT-mixing based model significantly surpasses that of ViT-based and ResNet-based ones. (ii)~Enhancing the feature dimension of the vision encoder results in notable improvements, e.g.,ResNet-34 + 6-layerViT vs.~ResNet-50 + 6-layerViT. (iii)~Increasing the number of shared encoder layers in the ViT-based model does not yield a significant enhancement in performance but incurs substantial computational overhead, {\em e.g.}, ResNet-50 + 6-layer ViT vs. ResNet-50 + 12-layer ViT.}

\change{Based on the comprehensive ablation results shown in Supplementary Section C, we choose the ResNet-ViT-mixing model with augmentation strategy and unifying the 3D scan depth to 32 as the most suitable settings for our following long-tailed case-level multi-modal diagnosis ability evaluation.} \change{After deciding the basic configurations, we evaluate the diagnosis ability on both our proposed internal \textbf{RP3D-DiagDS} dataset and various external benchmarks.
In internal evaluation, we mainly show the effectiveness of our newly proposed modules, considering the problem as a multi-label classification task under a long-tailed distribution. In external evaluation, we systematically investigate the boundary of our final model, serving as a diagnosis foundation model under both fine-tuning and zero-shot settings.}

\subsection*{Evaluation on Internal Benchmark}
\begin{figure}[htb!]
\includegraphics[width=\textwidth]{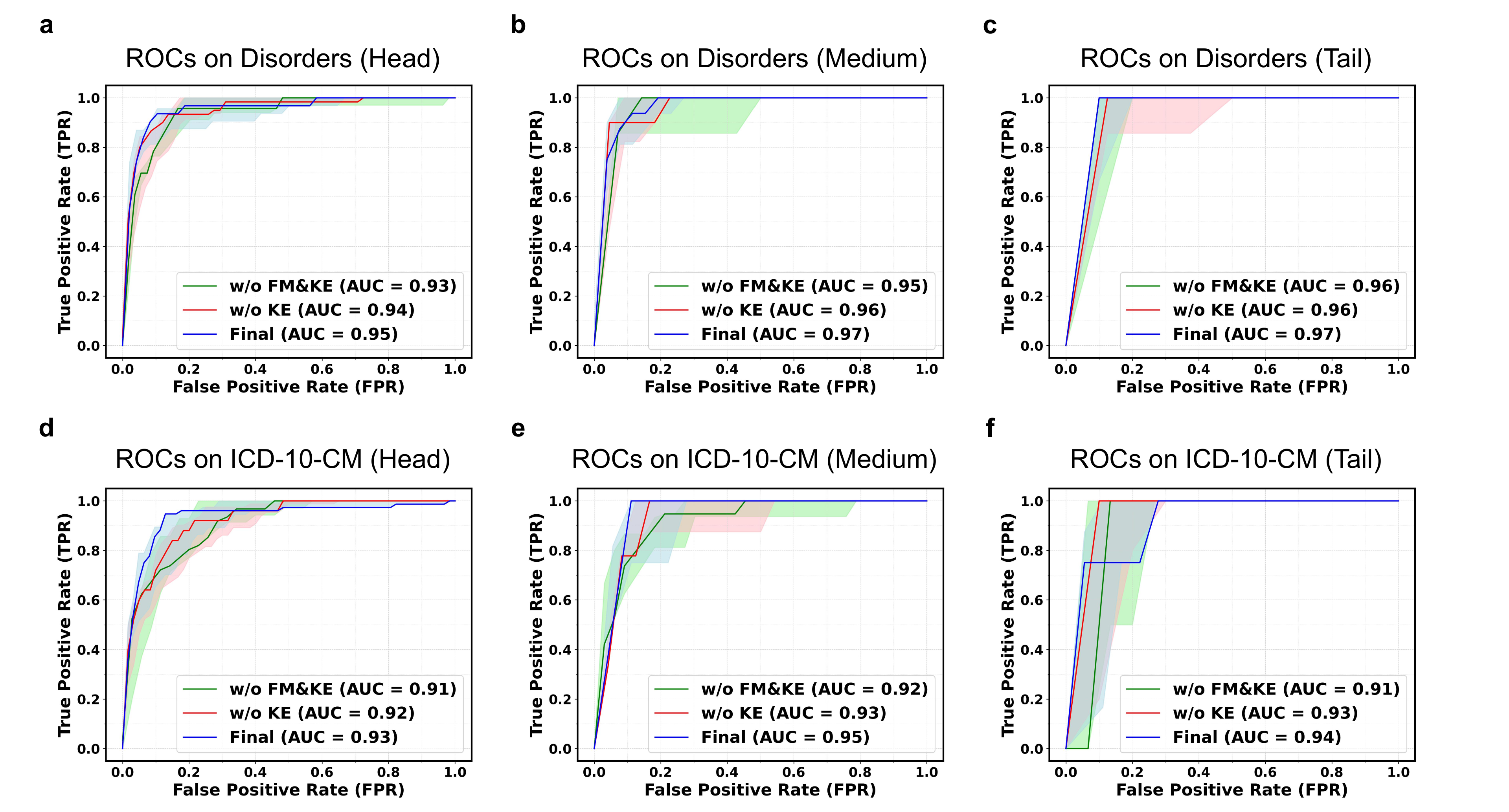}
\caption{\textbf{ROC curves on Disorders and ICD-10-CM, including head/medium/tail parts respectively.} The shadow in the figure shows the 95\% CI~(Confidence interval) and FM, KE is short for Fusion Module and Knowledge Enhancement. \change{More ROC curves for each class are shown in Supplementary Section F}}
\label{fig:ROCs Internal}
\end{figure}

In this part, our model is trained on the \textbf{entire} Rad3D-DiagDS training set~(29536 cases),
evaluated on its test split~(7772 cases), as detailed in Table~\ref{tab:knowledge_enhanced}. 
The comparison of AUC curves is depicted in Figure~\ref{fig:ROCs Internal}. We conduct experiments at two levels—disorder and ICD-10-CM classes—to evaluate the model's performance, both with and without text knowledge enhancement. 
These evaluations consider two integration strategies: early fusion and late fusion. For comprehensive definitions and applications of these fusion strategies, please see Supplementary Section B for more details. The model's effectiveness is assessed using various metrics on the test set, including the AUC curve's confidence interval (CI), to gauge its stability and consistency.

%In addition to demonstrate the superiority of our own methods, our evaluation results can also be viewed as a foundation benchmark for future works to compare with , further promoting the researches on developing better multi-modal multi-scan long-tailed diagnosis methods.
\begin{table}[!t]
\centering
\footnotesize 
\setlength{\tabcolsep}{7.5pt}
\caption{\change{\textbf{Classification results on Disorders and ICD-10-CM levels.} 
In the table ``FM'' represents the fusion module and ``KE'' represents the knowledge enhancement strategy. We report the results on Head/Medium/Tail \change{category} sets separately.}}
\label{tab:knowledge_enhanced}
\vspace{3pt}
\begin{tabular}{c|c|p{0.5cm}<{\centering}p{0.5cm}<{\centering}|ccccccc}
\toprule
\multirow{2}{*}{\textbf{Granularity}} & \multirow{2}{*}{\change{\textbf{Category}}} & \multicolumn{2}{c|}{\textbf{Methods}}                     & \multicolumn{7}{c}{\textbf{Metrics}}                                                                                 \\
                                      &                                    & FM                          & KE                          & \change{AUC}            & \change{AP}             & \change{F1}             & \change{MCC}            & \change{R@0.01}         & \change{F@0.05}         & \change{R@0.1}          \\ \midrule
\multirow{9}{*}{Disorders}            & \multirow{3}{*}{Head}              & \XSolidBrush & \XSolidBrush & 92.44          & 15.34          & 24.95          & 26.14          & 34.11          & 63.70          & 76.70          \\
                                      &                                    & \Checkmark   & \XSolidBrush & 93.61          & 18.16          & 28.15          & 29.30          & 39.89          & 67.74          & 80.10          \\
                                      &                                    & \Checkmark   & \Checkmark   & \textbf{94.41} & \textbf{20.27} & \textbf{30.21} & \textbf{32.27} & \textbf{41.93} & \textbf{71.09} & \textbf{81.15} \\ \cmidrule{2-11} 
                                      & \multirow{3}{*}{Medium}            & \XSolidBrush & \XSolidBrush & 93.74          & 11.12          & 19.88          & 22.67          & 32.62          & 62.29          & 73.63          \\
                                      &                                    & \Checkmark   & \XSolidBrush & 94.79          & 13.51          & 22.96          & 25.80          & 37.63          & 66.25          & 78.06          \\
                                      &                                    & \Checkmark   & \Checkmark   & \textbf{95.14} & \textbf{15.95} & \textbf{25.82} & \textbf{28.84} & \textbf{42.49} & \textbf{68.73} & \textbf{79.39} \\ \cmidrule{2-11} 
                                      & \multirow{3}{*}{Tail}              & \XSolidBrush & \XSolidBrush & 88.10          & 6.04           & 10.35          & 15.06          & 9.52           & 22.48          & 32.94          \\
                                      &                                    & \Checkmark   & \XSolidBrush & 90.13          & 6.65           & 10.98          & 15.84          & 11.43          & 27.37          & 43.01          \\
                                      &                                    & \Checkmark   & \Checkmark   & \textbf{90.96} & \textbf{7.75}  & \textbf{12.68} & \textbf{17.49} & \textbf{13.13} & \textbf{28.88} & \textbf{44.08} \\ \midrule
\multirow{9}{*}{ICD-10-CM}            & \multirow{3}{*}{Head}              & \XSolidBrush & \XSolidBrush & 89.83          & 12.31          & 20.27          & 21.58          & 23.81          & 52.05          & 68.45          \\
                                      &                                    & \Checkmark   & \XSolidBrush & 90.25          & 13.94          & 21.78          & 22.98          & 25.94          & 55.22          & 69.15          \\
                                      &                                    & \Checkmark   & \Checkmark   & \textbf{91.27} & \textbf{14.59} & \textbf{22.81} & \textbf{25.12} & \textbf{27.87} & \textbf{57.75} & \textbf{72.32} \\ \cmidrule{2-11} 
                                      & \multirow{3}{*}{Medium}            & \XSolidBrush & \XSolidBrush & 90.58          & 8.42           & 16.42          & 19.41          & 23.80          & 50.80          & 66.42          \\
                                      &                                    & \Checkmark   & \XSolidBrush & 91.12          & 9.18           & 18.02          & 20.05          & 27.10          & 53.77          & 67.45          \\
                                      &                                    & \Checkmark   & \Checkmark   & \textbf{92.01} & \textbf{10.34} & \textbf{19.08} & \textbf{22.16} & \textbf{28.81} & \textbf{55.55} & \textbf{69.86} \\ \cmidrule{2-11} 
                                      & \multirow{3}{*}{Tail}              & \XSolidBrush & \XSolidBrush & 86.91          & 4.82           & 8.64           & 12.70          & 10.13          & 29.82          & 51.04          \\
                                      &                                    & \Checkmark   & \XSolidBrush & 87.32          & 5.09           & 9.29           & 13.23          & 11.21          & 30.00          & 49.39          \\
                                      &                                    & \Checkmark   & \Checkmark   & \textbf{88.11} & \textbf{5.57}  & \textbf{10.48} & \textbf{14.68} & \textbf{12.75} & \textbf{30.11} & \textbf{51.77} \\ \bottomrule
\end{tabular}
\begin{tablenotes}
    \item [*]\change{*All indicators are presented as percentages, with the \% omitted in the table.}
    \end{tablenotes}  
\end{table}

% \begin{figure}[htb!]
%     \centering
%     \includegraphics[width=1\linewidth]{figure/section_Results/ROC_CI.pdf}
%     \caption{ROC curves on Disorders and ICD-10-CM, including head/medium/tail parts respectively. The shadow in the figure shown the 95\% CI~(Confidence interval) and FM, KE are short for Fusion Module and Knowledge Enhancement.}
%     \label{fig:knowledge_enhanced}
% \end{figure}

\noindent \textbf{Analysis on Fusion Module~(FM).} 
To demonstrate the effectiveness of the fusion module, we start with a baseline that adopts max pooling on the predictions of different images from the same case~(check Supplementary Section B for more details). Then, we add the fusion module and knowledge enhancement step-by-step, to improve the model's performance.

As shown in Table~\ref{tab:knowledge_enhanced}, 
adding the fusion module can greatly improve the results on Head, Medium, and Tail classes at both disorder and ICD-10-CM levels, showing the critical role of case-level information fusion in diagnosis tasks. These results align well with our expectations, as in clinical practice, the examination of one modality for a diagnosis is often insufficient. A thorough and meticulous diagnostic process typically involves an integrated review of all test results. Each test is weighted differently, depending on how its results correspond with other tests. Our fusion model adeptly mirrors this comprehensive approach, demonstrating its effectiveness in simulating the nuanced process of clinical diagnosis.

%and former models that only enable analysis on a single scan can hardly meet the demand.  Our fusion module can also meet the clinical practical demand where patients may obtain additional radiologic examination results at any time with the expectation of a more accurate diagnosis.

\noindent \textbf{Analysis on Knowledge Enhancement~(KE).} 
We experiment with the knowledge-enhanced training strategy, 
where we first leverage the rich domain knowledge to pre-train a knowledge encoder with natural language and use it to guide the visual representation learning for disease diagnosis. As shown in Table~\ref{tab:knowledge_enhanced}, 
knowledge enhancement further promotes the final diagnosis performance on both disorder and ICD-10-CM classifications, showing that a better text embedding space trained by rich domain knowledge can greatly help visual representation learning.

\noindent \textbf{Analysis on ROC Curves.} 
The solid lines in the Figure~\ref{fig:ROCs Internal} represents the median AUC value. This value is derived from a process where, for each category, 1000 random samples are taken and the median AUC value is calculated. This procedure is repeated multiple times (1000), with the final curve representing the median of these values. 
The ROC curve is illustrated by the solid lines. 
Accompanying each solid line is a shaded area, 
which denotes the 95\% confidence interval (CI). 
For each type of implementation, 
this shaded area is significantly higher than the portion above the solid line. This implies that the AUC value represented by the solid line is generally higher than the class-wise average across different data splits. As shown in the figure, the highest AUC value has reached 0.97, and the lowest value is 0.91, this observed pattern suggests, that there exists a few categories that are more challenging to learn than others.

\change{\noindent \textbf{Analysis on Normal/Abnormal Diagnosis.}
In clinical scenarios, the proportion of normal cases is significantly higher compared to abnormal cases. Therefore, it is crucial to highlight the evaluation of the model's ability to distinguish between normal and abnormal cases. 
As shown in Figure~\ref{fig:NormalAnalysis}a, 
our model achieves an AUC of 82.61\% for such diagnosis, 
given the significant disease variations in abnormal cases. Additionally, we also split out six test sub-sets based on the anatomy of cases, including Head and Neck, Chest, Abdomen and Pelvis, Upper Limb, Lower Limb, Spine, and Breast. In each sub-set, all cases are from the same anatomy, thus we can track the performance among a certain anatomy region. As shown in Figure~\ref{fig:NormalAnalysis}a, 
the results on each anatomy are generally consistent, indicating that
our model robustly demonstrates superior performance across different
anatomical regions, without showing notable biases on any specific area.}

% \change{In clinical scenarios, the proportion of normal cases is significantly higher compared to abnormal cases. Therefore, it is crucial to separately evaluate the model's ability to distinguish between normal and abnormal cases. Simultaneously, it is precisely because the normal class is quite common and exists in a large number of various clinical data sets, to further enhance our model's performance on the normal class, we finetune it using normal cases from additional external datasets~(\textbf{these datasets will not overlap with the data sets we will test below}). These additional datasets include Brain-Tumor-17, CheXpert, Vertebrase-Xray, KneeMRI, and VinDr-Mammo. We design two sets of experiments to assess this capability using both internal data from RP3D-DiagDS and external clinical data. For reference and comparison, we test RadImageNet (\textbf{RadIN}), \textbf{BiomedCLIP}, and our model after finetune with additional normal data (\textbf{Ours}).}

\begin{figure}[htb!]
\includegraphics[width=\textwidth]{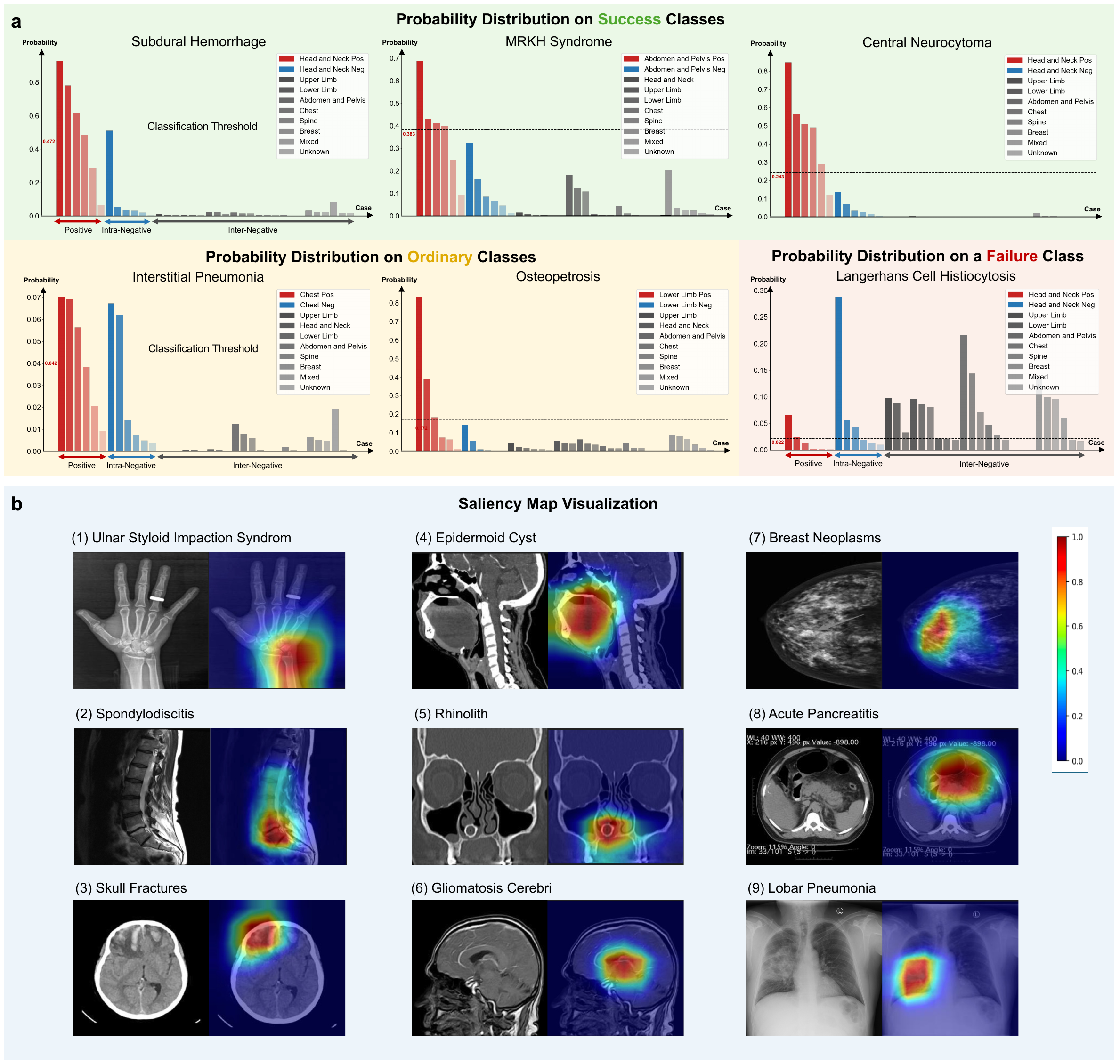}
\caption{
\change{\textbf{The prediction probability distribution for different classes and saliency map visualization.} \textbf{a} Based on the anatomies of each case, we split them into positive cases, intra-negative cases which are located in the same anatomy as the positive ones, and inter-negative cases which are located in other anatomies. The classification threshold score in the figure denotes the final comparison bar to transform the soft probabilities into binary true/false diagnosis results. The first three probability distribution figures depict the distributions of three relatively successful classes, where the model can clearly distinguish the inter-negative cases and the intra-negative cases are more confusing.
 We then show two ordinary classes. As shown by the distributions, most errors are caused by the intra-negative cases, and similarly, the inter-negative cases are easily dismissed as well.
 At last, we show a failure case where the model can hardly distinguish the positive and negative cases regardless of whether they are intra-negative or inter-negative.
\textbf{b} Saliency map of the key frames. Red indicates the areas that the model focuses on when inferring the corresponding disease category. This indicates that \textbf{RadDiag} is capable of accurately identifying the locations of lesions or abnormal regions. 
}}
\label{fig:ContrastVisualization}
\end{figure}

\noindent \textbf{\change{Analysis on Prediction Probability Distribution.}}
\change{To get a deeper and more intuitive assessment of the model's capabilities, we analyzed its performance across different disease classes. Given the large number of diseases, we pick a total of six diseases, as shown in Figure~\ref{fig:ContrastVisualization},
where we demonstrate the prediction probability distribution on cases in the test set for a certain class. 
All cases are categorized by their related anatomies, with positive cases corresponding to the considered disease class and negative cases divided into two types: intra-negative cases from the same anatomy, and inter-negative cases from other anatomies. 
Horizontal dashed lines in the bar plot denotes the threshold, 
with positive and negative cases from the same anatomy colored in \textbf{\textcolor{red}{red}} and \textbf{\textcolor{blue}{blue}}, and negative cases from other anatomies in different shades of \textbf{\textcolor{gray}{gray}}.
Ideally, positive cases should have high scores, while both types of negative cases should have low scores. Therefore, the difference between the actual distribution and the ideal situation can effectively illustrate the model's performance on the disease. 
As shown by the results, for the success classes, the model can distinguish both the intra-negative and inter-negative cases against the positive, while for the ordinary classes, the model may be confused with some intra-negative and positive cases. 
In the failure class, the model cannot pick out the positive cases.
These results demonstrate that our internal evaluation requires the model to perform a two-stage differential diagnosis. 
The first step is to locate the possible diseases through the shooting anatomy, {\em etc.}, and then capture the detailed lesion patterns features to make further diagnosis.}

\noindent \textbf{\change{Analysis on Saliency Map Visualization.}} \change{Additionally, inspired by the proposal of explainable AI~(XAI)~\cite{joyce2023explainable}, we utilize Score-CAM~\cite{wang2020score}, a visual explanation method based on class activation maps, to visualize the areas that contribute more to make a decision on the target disease class. 
As shown in Figure~\ref{fig:ContrastVisualization}b, we sample several cases from different modalities and anatomies to compute the saliency map. For instance, as depicted in Figure~\ref{fig:ContrastVisualization}b(1) for ``Ulnar styloid impaction syndrome'', our model can pinpoint the location of the ulna during inference for this category, without being distracted by other anomalies, such as rings worn on the four finger. 
In another case in Figure~\ref{fig:ContrastVisualization}b(2), for the category of ``Spondylodiscitis'', our model shows its ability to focus on the intervertebral disk, with heightened attention particularly on the region of abnormality.
It proves our model's precision in identifying the relevant anatomical structures for the disease categories, even amidst other potential distractions or anomalies within the same image, thus illustrating its robustness and specificity in medical disease diagnosis.} 

% \begin{figure}[!ht]
%     \centering
%     \includegraphics[width=1\linewidth]{figure/section_Results/cam_3.pdf}
%     \caption{\change{Saliency map of the key frames. Red indicates the areas that our model focuses on when inferring the corresponding disease category. It indicates that our model is capable of accurately identifying the locations of lesions or abnormal regions. }}
%     \label{fig:cam}
% \end{figure}

% \change{\noindent \textbf{Inter-slice Attention Explanation.} In our adopted ResNet + ViT model architecture, we can assess the model's attention to each slice through the attention map weights in the ViT structure.}

% \change{In these cases from Radiopaedia, the preset images of 3D scans are typically selected by physicians as representative slices that best reflect the pathological features of the disease in question. Consequently, we consider these slices as key frames within the entire 3D volume. It is important to note that though disease diagnosis cannot be performed only through key frames or only through key frames, the two have a strong positive correlation. Therefore, to evaluate the effectiveness and interpretability of the attention mechanism in our model, we conduct a zero-shot experiment to assess the correlation between the attention maps and the key slices, without incorporating key-slice supervision during training. When we consider that the five frames around the key slice can be considered as relatively key slices, the model can achieve 85.04\% recognition accuracy for key slices.}

\subsection*{Evaluation on External Benchmarks}

\change{In this part, we further assess the transferring ability of our diagnosis model trained on \textbf{RP3D-DiagDS}. 
We adopt two settings to demonstrate the transferring ability to External Benchmarks of our model, \emph{i.e.}, zero-shot, and fine-tuning. \textit{First}, in the zero-shot setting, we evaluate the effectiveness of our model for directly deployed in real clinical scenarios. In detail, as our dataset encompasses a wide range of disease categories affecting the entire body, it can be tested on most external datasets.
More details on how we perform zero-shot transferring for our method and other baselines can be found in Section~\ref{baselines}. We have adopted F1, MCC, and ACC for comparison. }
\textit{Second}, our model can also serve as a pre-trained model and be fine-tuned on each downstream task to improve the final performance. Specifically, for the dataset with single image input, we simply adopt the pre-trained visual encoder module, 
{\em i.e.}, discarding the fusion module. For the dataset with multi-image input, 
we inherit the entire model. In both cases, the final classification layer will be trained from scratch. In addition to using all available external training data,  
we also consider using 1\%, 10\%, and 30\% portion data for few-shot learning.

\begin{figure}[!t]
    \centering
    \includegraphics[width=1\linewidth]{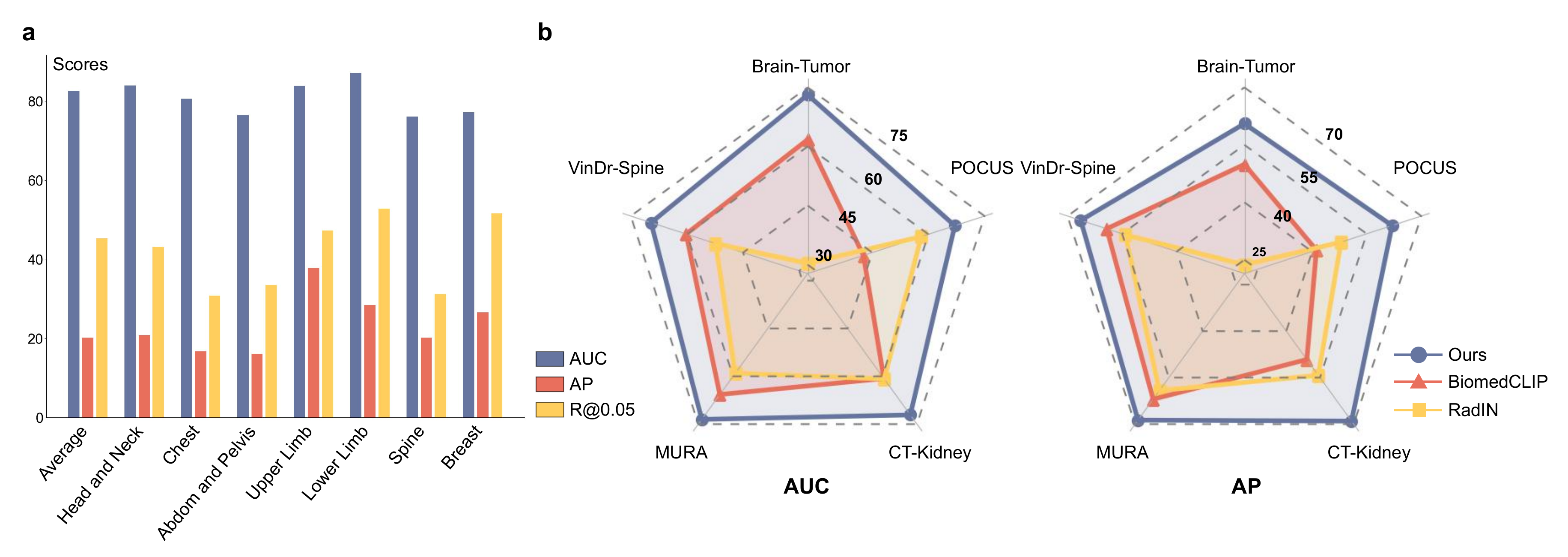}
    \caption{\change{\textbf{Analysis on normal/abnormal diagnosis.} \textbf{a} The bar plot shows the AUC, AP, and Recall@0.05 of the Normal class on different body regions in the Internal test set of RP3D-DiagDS. \textbf{b} The radar figure shows the AUC and AP performance comparison between \textbf{RadIN}, \textbf{BiomedCLIP}, and \textbf{Ours} on external clinical datasets at five different anatomies.}}
    \label{fig:NormalAnalysis}
\end{figure}

\noindent \textbf{\change{Zero-shot Results on Normal/Abnormal Diagnosis.}} 
\change{
In zero-shot assessment, we evaluate the transferring ability of our final model on \textbf{normal/abnormal diagnosis} for external dataset, \emph{e.g.}, Brain-Tumor (405 normal, 906 abnormal), POCUS (256 normal, 302 abnormal), CT-Kidney (1037 normal, 1453 abnormal), MURA (1667 normal, 1530 abnormal), and VinDr-Spine (1070 normal, 1007 abnormal), totaling 9634 cases. These datasets are derived from medical centers in different regions, thus is a good assessment of the model's performance in distinguishing normal from abnormal cases under zero-shot conditions.  
We directly adopt the ``normal'' classifier head trained on our \textbf{RP3D-DiagDS} to obtain the prediction probability score and compare it with other foundation models, RadIN~\cite{mei2022radimagenet} and BiomedCLIP~\cite{zhang2023biomedclip},
in terms of AUC and AP on the five zero-shot datasets. 
As shown in Figure~\ref{fig:NormalAnalysis}b, our model demonstrates significantly better results on all the five considered datasets, \emph{i.e.}, Brain-Tumor, POCUS, CT-Kidney, and MURA, compared to RadIN and BiomedCLIP. 
These findings show that our model consistently demonstrates superior diagnostic performance on zero-shot normal/abnormal diagnosis, across different anatomies under cross-center real clinical practices, significantly outperforming existing SOTA models.}

\noindent \textbf{\change{Zero-shot Results on Fine-grained Diagnosis.}}
\change{In addition to the normal/abnormal diagnosis, as shown in Table~\ref{tab:External Zeroshot}, we also conduct zero-shot evaluation on \textbf{fine-grained diagnosis tasks} introduced by each representative external datasets, spanning six anatomies, including the head and neck, chest, breast, abdomen and pelvis, and spine. These datasets cover five modalities, \emph{i.e.}, CT, MRI, X-ray, Ultrasound, and Mammography, allowing for a comprehensive evaluation of our model's capabilities. As baselines, we have included RadIN~\cite{mei2022radimagenet} and BiomedCLIP~\cite{zhang2023biomedclip} for comparison in the zero-shot setting. The results demonstrate that our model outperforms the other two models in overall performance, indicating strong transfer learning capabilities. Interestingly, our model's zero-shot performance can sometimes be even comparable to the results obtained by the model trained from scratch on a subset of the corresponding dataset, demonstrating the superiority of our RadDiag for diagnostic capabilities, which cannot be achieved by former foundation models.}

\begin{table}[!t]
\begin{threeparttable}
    \centering
    \footnotesize
    \setlength{\tabcolsep}{7.5pt}
\caption{\change{\textbf{Zero-shot results on 6 external datasets.} For our model, we avoid exposing to these external data for zero-shot experiments and use the threshold corresponding to the highest F1 score on the evaluation set. For RadIN and BiomedCLIP, we use the threshold corresponding to the highest F1 score on each individual external dataset. We also report the supervised training results with partially down-stream task external data, denoting as ``S.T.'', for demonstrating what diagnosis performance our model can achieve zero-shotly. We have evaluated F1, MCC and ACC, and all the results reported in the table are the averaged values across all classes. In the table, "US" represents ultrasound and "MG" represents mammography.}}
\begin{tabular}{llll|l|c|ccc}
\toprule
Dataset                                          & Anatomy                           & Modality                   & Dim                    &  Method  & External Data     & F1 $\uparrow$      & MCC $\uparrow$     & ACC $\uparrow$  \\ \midrule
\multicolumn{1}{c}{\multirow{4}{*}{Brain-Tumor}} & \multirow{4}{*}{Head\&Neck}    & \multirow{4}{*}{MRI}       & \multirow{4}{*}{2D}    & RadIN       & \multirow{3}{*}{Zero-shot} & 22.29 & 3.23  & 63.63 \\ 
\multicolumn{1}{c}{}                             &                                   &                            &                        & BiomedCLIP  &  & 56.25 & 32.81 & 57.21 \\ 
\multicolumn{1}{c}{}                             &                                   &                            &                        & Ours        &  & 61.09 & 47.37 & 73.38 \\ \cmidrule{5-9}
                                                 &                                   &                            &                        & \cellcolor[gray]{0.85}S.T. & \cellcolor[gray]{0.85}n=571(10\%)  &\cellcolor[gray]{0.85}61.77 &\cellcolor[gray]{0.85}46.69 &\cellcolor[gray]{0.85}72.17 \\\midrule
\multirow{4}{*}{POCUS}                           & \multirow{4}{*}{Chest}            & \multirow{4}{*}{US}        & \multirow{4}{*}{Video} & RadIN       & \multirow{3}{*}{Zero-shot} & 2.22  & 0.17  & 66.67 \\ 
                                                 &                                   &                            &                        & BiomedCLIP   &  &50.33 & 11.73 & 45.82 \\ 
                                                 &                                   &                            &                        & Ours       &  & 65.02 & 47.97 & 79.84 \\ \cmidrule{5-9}
                                                 &                                   &                            &                        & \cellcolor[gray]{0.85}S.T. & \cellcolor[gray]{0.85}n=223(1\%)  & \cellcolor[gray]{0.85}64.20 & \cellcolor[gray]{0.85}48.31 & \cellcolor[gray]{0.85}80.58 \\\midrule
\multirow{4}{*}{CT-KIDNEY}                       & \multirow{4}{*}{Abdom\&Pelvis} & \multirow{4}{*}{CT}        & \multirow{4}{*}{2D}    & RadIN      & \multirow{3}{*}{Zero-shot} & 38.1  & 17.57 & 64.75 \\ 
                                                 &                                   &                            &                        & BiomedCLIP   & & 49.38 & 29.58 & 72.21 \\ 
                                                 &                                   &                            &                        & Ours        &  & 56.50 & 43.18 & 69.96 \\ \cmidrule{5-9}
                                                 &                                   &                            &                        & \cellcolor[gray]{0.85}S.T. & \cellcolor[gray]{0.85}n=995(10\%)      & \cellcolor[gray]{0.85}55.73 & \cellcolor[gray]{0.85}42.28 & \cellcolor[gray]{0.85}66.95 \\\midrule
\multirow{4}{*}{MURA}                            & \multirow{4}{*}{Limb}             & \multirow{4}{*}{X-ray}     & \multirow{4}{*}{2D}    & RadIN       & \multirow{3}{*}{Zero-shot} & 31.26 & 4.37  & 52.94 \\ 
                                                 &                                   &                            &                        & BiomedCLIP  &  & 68.36 & 14.96 & 54.70 \\ 
                                                 &                                   &                            &                        & Ours        &  & 67.64 & 40.15 & 77.51 \\\cmidrule{5-9}
                                                 &                                   &                            &                        &\cellcolor[gray]{0.85}S.T. & \cellcolor[gray]{0.85}n=3680(10\%)    &\cellcolor[gray]{0.85}66.71 &\cellcolor[gray]{0.85}41.98 &\cellcolor[gray]{0.85}78.32 \\ \midrule
\multirow{4}{*}{VinDr-Spine}                     & \multirow{4}{*}{Spine}            & \multirow{4}{*}{X-ray}     & \multirow{4}{*}{2D}    & RadIN       & \multirow{3}{*}{Zero-shot} & 4.34  & 0.91  & 87.49 \\ 
                                                 &                                   &                            &                        & BiomedCLIP  &  & 26.49 & 12.79 & 61.3  \\ 
                                                 &                                   &                            &                        & Ours        &  & 25.93 & 19.40 & 83.07 \\ \cmidrule{5-9}
                                                 &                                   &                            &                        &\cellcolor[gray]{0.85}S.T. & \cellcolor[gray]{0.85}n=84(1\%)      & \cellcolor[gray]{0.85}24.73 & \cellcolor[gray]{0.85}20.61 & \cellcolor[gray]{0.85}84.46 \\\midrule
\multirow{4}{*}{TCGA}                            & \multirow{4}{*}{Chest\&Breast} & \multirow{4}{*}{CT\&MG} & \multirow{4}{*}{3D}    & RadIN          & \multirow{3}{*}{Zero-shot} & -     & -     & -     \\ 
                                                 &                                   &                            &                        & BiomedCLIP     &  & -     & -     & -     \\ 
                                                 &                                   &                            &                        & Ours        &  & 72.18 & 57.20 & 83.47 \\ \cmidrule{5-9}
                                                 &                                   &                            &                        & \cellcolor[gray]{0.85}S.T. & \cellcolor[gray]{0.85}n=173(30\%)      & \cellcolor[gray]{0.85}70.33 & \cellcolor[gray]{0.85}52.26 & \cellcolor[gray]{0.85}77.41 \\
                                                 \bottomrule
\end{tabular}
\begin{tablenotes}
    \item [*] \change{All indicators are presented as percentages, with the \% omitted in the table.}
    \end{tablenotes}   
    \label{tab:External Zeroshot}
\end{threeparttable} 
\end{table}

% \change{\textbf{Internal RP3D-DiagDS Data.} We classify all normal cases from various anatomies as a distinct category for model prediction. In the entire dataset, this category has a higher frequency than any other individual disease category.  Additionally, we selected over 200 normal cases (covering different anatomical regions) for further evaluation of Head and Neck, Chest, Abdomen and Pelvis, Upper Limb, Lower Limb, Spine, and Breast. To ensure the robustness of our results, we evaluate normal cases of each anatomy by mixing them only with other cases from the same anatomy.}

\noindent \textbf{\change{Finetuning Results.}} 
As shown in Table~\ref{tab:External Evaluation}, 
while finetuning our model on external datasets, 
we see significant performance improvement on all 22 external datasets with different data portions, compared to models trained from scratch. 
In most cases, our model can even surpass the specialist SOTAs which are designed carefully for the targeting task, 
demonstrating that publicly shared medical data on the Internet is a tremendous and valuable resource that can serve as a superior large-scale supervised training dataset for the medical domain. 

\begin{table}[!ht]
\begin{threeparttable}
    \centering
    \footnotesize
    \setlength{\tabcolsep}{4pt}
    \caption{\textbf{The AUC score comparison on various external datasets.} For each dataset, we experimented with different training data portions, denoted as 1\% to 100\% in the table. For example, 30\% represents we use 30\% of data in the downstream training set for finetuning our model or training from scratch. ``SOTA'' denotes the best performance of former works~(pointed with corresponding reference) on the datasets.  We mark the gap between ours and training from scratch on the subscript of \textcolor{darkpastelgreen}{\textbf{uparrows$\uparrow$}} in the table.}
    % \xiaoman{why not introduce NSCLC< TCGA, ISPY1 in the datasets part.}\qiaoyu{qiaoyu: fixed}}
    \centering
    \begin{tabular}{l|cc|cc|cc|cc|c}
    \toprule
    \multirow{2}{*}{\textbf{Dataset}}  & \multicolumn{2}{c|}{1\%} & \multicolumn{2}{c|}{10\%} & \multicolumn{2}{c|}{30\%} & \multicolumn{2}{c|}{100\%} & \multirow{2}{*}{SOTA}\\        
                              & \textbf{Scratch} & \textbf{Ours}   & \textbf{Scratch} & \textbf{Ours}  & \textbf{Scratch} & \textbf{Ours}  & \textbf{Scratch} & \textbf{Ours} &       \\ \midrule
    VinDr-Mammo                 &   57.05  &   \textbf{58.69}\textcolor{darkpastelgreen}{$\uparrow^{1.64}$}    &   58.22 & \textbf{59.55}\textcolor{darkpastelgreen}{$\uparrow^{1.33}$} & 62.10 & \textbf{63.04}\textcolor{darkpastelgreen}{$\uparrow^{0.94}$} & 76.25 &   \textbf{78.46}\textcolor{darkpastelgreen}{$\uparrow^{2.21}$} & 77.50\textsuperscript{\dag}~\cite{bhat2023aucreshaping} \\ 
    CXR14                             &    76.85   &   \textbf{78.87}\textcolor{darkpastelgreen}{$\uparrow^{2.02}$} &     77.93      &  \textbf{80.66}\textcolor{darkpastelgreen}{$\uparrow^{2.73}$}         &  78.52    & \textbf{81.22}\textcolor{darkpastelgreen}{$\uparrow^{2.70}$}  &    79.12     &  \textbf{83.44}\textcolor{darkpastelgreen}{$\uparrow^{4.32}$}  &   82.50\textsuperscript{\dag}~\cite{zhang2023knowledge}           \\ 
    VinDr-Spine     &      79.35         &    \textbf{82.13}\textcolor{darkpastelgreen}{$\uparrow^{2.78}$}   &      85.02        &   \textbf{86.43}\textcolor{darkpastelgreen}{$\uparrow^{1.41}$}         &    86.90  & \textbf{87.33}\textcolor{darkpastelgreen}{$\uparrow^{0.43}$}   &    87.35        &     87.92\textcolor{darkpastelgreen}{$\uparrow^{0.57}$}  &    \textbf{88.90}\textsuperscript{*}~\cite{Wu2023KDiagKD}   \\ 
    MosMedData & 52.96 & \textbf{60.78}\textcolor{darkpastelgreen}{$\uparrow^{7.82}$} & 62.11 & \textbf{64.66}\textcolor{darkpastelgreen}{$\uparrow^{2.55}$} & 65.63 & \textbf{70.96}\textcolor{darkpastelgreen}{$\uparrow^{5.33}$} & 72.36 & \textbf{76.79}\textcolor{darkpastelgreen}{$\uparrow^{4.43}$} & 68.47\textsuperscript{\dag}~\cite{morozov2020mosmeddata}  \\ 
    ADNI & 56.47 & \textbf{59.19}\textcolor{darkpastelgreen}{$\uparrow^{2.72}$} & 61.62 & \textbf{65.08}\textcolor{darkpastelgreen}{$\uparrow^{3.46}$} & 65.13 & \textbf{66.98}\textcolor{darkpastelgreen}{$\uparrow^{1.85}$} & 83.44  & \textbf{85.61}\textcolor{darkpastelgreen}{$\uparrow^{2.17}$} & 79.34\textsuperscript{\dag}~\cite{korolev2017residual} \\ 
    NSCLC & 53.19 & \textbf{58.00}\textcolor{darkpastelgreen}{$\uparrow^{4.81}$} & 55.21 & \textbf{59.37}\textcolor{darkpastelgreen}{$\uparrow^{4.16}$} & 57.33 & \textbf{63.01}\textcolor{darkpastelgreen}{$\uparrow^{5.68}$} & 67.25 & \textbf{72.54}\textcolor{darkpastelgreen}{$\uparrow^{5.29}$} & N/A \\ 
    TCGA & 55.28 &  \textbf{67.33}\textcolor{darkpastelgreen}{$\uparrow^{12.05}$} & 71.16 & \textbf{76.91}\textcolor{darkpastelgreen}{$\uparrow^{5.75}$} & 78.55 & \textbf{85.27}\textcolor{darkpastelgreen}{$\uparrow^{6.72}$} & 88.66 & \textbf{95.17}\textcolor{darkpastelgreen}{$\uparrow^{6.51}$} & N/A \\ 
    ISPY1 & 52.61 &  \textbf{57.99}\textcolor{darkpastelgreen}{$\uparrow^{5.38}$} & 57.32 & \textbf{59.89}\textcolor{darkpastelgreen}{$\uparrow^{2.57}$} & 58.13 & \textbf{62.22}\textcolor{darkpastelgreen}{$\uparrow^{4.09}$} & 65.88 & \textbf{69.43}\textcolor{darkpastelgreen}{$\uparrow^{3.55}$} & N/A \\ 
    \change{ChestX-Det10} & 60.12 & \textbf{64.40}\textcolor{darkpastelgreen}{$\uparrow^{4.28}$} & 64.55 & \textbf{67.61}\textcolor{darkpastelgreen}{$\uparrow^{3.06}$} & 71.38 & \textbf{75.27}\textcolor{darkpastelgreen}{$\uparrow^{3.89}$} & 74.11 & \textbf{79.44}\textcolor{darkpastelgreen}{$\uparrow^{5.33}$} & N/A \\ 
    
    \change{CheXpert} & 85.14 & \textbf{87.55}\textcolor{darkpastelgreen}{$\uparrow^{2.41}$} & 87.97 & \textbf{89.26}\textcolor{darkpastelgreen}{$\uparrow^{1.29}$} & 88.31 & \textbf{90.08}\textcolor{darkpastelgreen}{$\uparrow^{1.77}$} & 89.52 & 91.27\textcolor{darkpastelgreen}{$\uparrow^{1.75}$} &  \textbf{93.00}\textsuperscript{*}~\cite{zhang2023knowledge} \\ 
    \change{COVID-19-Radio} & 86.67 & \textbf{88.34}\textcolor{darkpastelgreen}{$\uparrow^{1.67}$} & 87.47 & \textbf{90.38}\textcolor{darkpastelgreen}{$\uparrow^{2.91}$} & 90.66 & \textbf{92.71}\textcolor{darkpastelgreen}{$\uparrow^{2.05}$} & 95.39 & \textbf{98.56}\textcolor{darkpastelgreen}{$\uparrow^{3.17}$} & N/A \\  
    \change{IU-Xray} & 60.22 & \textbf{63.38}\textcolor{darkpastelgreen}{$\uparrow^{3.16}$} & 63.74 & \textbf{66.39}\textcolor{darkpastelgreen}{$\uparrow^{2.65}$} & 67.55 & \textbf{70.27}\textcolor{darkpastelgreen}{$\uparrow^{2.72}$} & 74.01 & \textbf{76.04}\textcolor{darkpastelgreen}{$\uparrow^{2.03}$} & N/A \\  
    \change{LNDb} & 55.00 & \textbf{57.21}\textcolor{darkpastelgreen}{$\uparrow^{2.21}$} & 60.77 & \textbf{62.10}\textcolor{darkpastelgreen}{$\uparrow^{1.33}$} & 63.55 & \textbf{64.31}\textcolor{darkpastelgreen}{$\uparrow^{0.76}$} & 68.76 & \textbf{70.58}\textcolor{darkpastelgreen}{$\uparrow^{1.82}$} & N/A \\ 
    \change{PadChest} & 77.10 & \textbf{78.36}\textcolor{darkpastelgreen}{$\uparrow^{1.25}$} & 81.21 & \textbf{82.31}\textcolor{darkpastelgreen}{$\uparrow^{1.10}$} & 82.45 & \textbf{84.18}\textcolor{darkpastelgreen}{$\uparrow^{1.73}$} & 83.66 & \textbf{85.87}\textcolor{darkpastelgreen}{$\uparrow^{2.21}$} & N/A \\  
    \change{CC-CCII} & 86.55 & 89.30\textcolor{darkpastelgreen}{$\uparrow^{2.75}$} & 94.19 & 96.30\textcolor{darkpastelgreen}{$\uparrow^{2.11}$} & 95.04 & 97.81\textcolor{darkpastelgreen}{$\uparrow^{2.77}$} & 98.27 & \textbf{99.46}\textcolor{darkpastelgreen}{$\uparrow^{1.19}$} & 97.41\textsuperscript{\dag}~\cite{zhang2020clinically} \\ 
    \change{RadChest} & 59.92 & \textbf{61.36}\textcolor{darkpastelgreen}{$\uparrow^{1.44}$} & 66.11 & \textbf{67.36}\textcolor{darkpastelgreen}{$\uparrow^{1.25}$} & 70.09 & \textbf{72.18}\textcolor{darkpastelgreen}{$\uparrow^{2.09}$} & 74.22 & 76.15\textcolor{darkpastelgreen}{$\uparrow^{1.93}$} & \textbf{77.30}\textsuperscript{*}~\cite{draelos2021machine} \\  
    \change{Brain-Tumor} & 75.61 & \textbf{77.14}\textcolor{darkpastelgreen}{$\uparrow^{1.53}$} & 80.86 & \textbf{82.33}\textcolor{darkpastelgreen}{$\uparrow^{1.47}$} & 84.57 & \textbf{88.77}\textcolor{darkpastelgreen}{$\uparrow^{4.20}$} & 91.05 & \textbf{93.21}\textcolor{darkpastelgreen}{$\uparrow^{2.16}$} & N/A \\  
    \change{Brain-Tumor-17} & 69.98 & \textbf{72.42}\textcolor{darkpastelgreen}{$\uparrow^{2.44}$} & 78.78 & \textbf{80.31}\textcolor{darkpastelgreen}{$\uparrow^{1.53}$} & 87.49 & \textbf{91.01}\textcolor{darkpastelgreen}{$\uparrow^{3.52}$} & 93.66 & \textbf{94.43}\textcolor{darkpastelgreen}{$\uparrow^{0.77}$} & N/A \\ 
    \change{POCUS} & 78.20 & \textbf{79.34}\textcolor{darkpastelgreen}{$\uparrow^{1.14}$} & 84.09 & \textbf{85.76}\textcolor{darkpastelgreen}{$\uparrow^{1.67}$} & 88.42 & \textbf{89.66}\textcolor{darkpastelgreen}{$\uparrow^{1.24}$} & 94.31 & \textbf{95.46}\textcolor{darkpastelgreen}{$\uparrow^{1.15}$} & 94.00\textsuperscript{\dag}~\cite{born2021accelerating} \\  
   \change{MURA} & 76.28 & \textbf{78.19}\textcolor{darkpastelgreen}{$\uparrow^{1.91}$} & 81.41 & \textbf{84.33}\textcolor{darkpastelgreen}{$\uparrow^{2.92}$} & 85.17 & \textbf{86.27}\textcolor{darkpastelgreen}{$\uparrow^{1.10}$} & 86.25 & 88.31\textcolor{darkpastelgreen}{$\uparrow^{2.06}$} & \textbf{92.90}\textsuperscript{*}~\cite{rajpurkar2017mura} \\ 
    \change{KneeMRI} & 55.06 & \textbf{58.88}\textcolor{darkpastelgreen}{$\uparrow^{3.82}$} & 62.88 & \textbf{63.12}\textcolor{darkpastelgreen}{$\uparrow^{0.24}$} & 67.19 & \textbf{68.36}\textcolor{darkpastelgreen}{$\uparrow^{1.17}$} & 73.05 & \textbf{73.22}\textcolor{darkpastelgreen}{$\uparrow^{0.17}$} & N/A \\ 
    \change{CT-Kidney} & 76.10 & \textbf{79.17}\textcolor{darkpastelgreen}{$\uparrow^{3.07}$} & 83.61 & \textbf{84.96}\textcolor{darkpastelgreen}{$\uparrow^{1.35}$} & 85.99 & \textbf{88.22}\textcolor{darkpastelgreen}{$\uparrow^{2.23}$} & 91.41 & \textbf{93.26}\textcolor{darkpastelgreen}{$\uparrow^{1.85}$} & N/A \\ 
    \bottomrule
    \end{tabular}
    \begin{tablenotes}
    \item[*] The numbers are borrowed from the referred papers. 
    \item[\dag] \change{The four SOTAs~\cite{morozov2020mosmeddata,korolev2017residual,zhang2020clinically,born2021accelerating} are relatively old. Thus, their network architecture is weaker than the classical 3D ResNet trained from scratch.}
    \item [-]\change{All indicators are presented as percentages, with the \% omitted in the table.}
    \end{tablenotes}   
    \label{tab:External Evaluation}
\end{threeparttable}
\end{table}

\section{DISCUSSION}

There have been a number of open-source datasets for disease diagnosis. Notably, large-scale datasets such as NIH ChestX-ray~\cite{wang2017chestx}, MIMIC-CXR~\cite{johnson2019mimic}, and CheXpert~\cite{irvin2019chexpert} stand out for their comprehensive collection of annotated X-ray images, facilitating extensive research and advancements in automated disease classification.
However, it's important to note three key limitations.
First, these open-source diagnosis datasets mainly consist of chest X-rays~\cite{Nguyen2022VinDrPCXRAO,bustos2020padchest,wang2017chestx,irvin2019chexpert,majkowska2020chest,objectcxr,jaeger2014two,shih2019augmenting,filice2020crowdsourcing,nguyen2022vindr,johnson2019mimic, chowdhury2020can}, thus are 2D images. 
There are only a few 3D datasets available, with a limited number of volumes~\cite{msoud_nickparvar_2021,SHREYA_2023,FERNANDO_2023,KOENIG2020102248,Marcus2007,data5010014,born2020pocovid,armato2011lung,bien2018deep}.
Second, a significant portion of these large-scale datasets focuses solely on binary classification of specific diseases~\cite{shih2019augmenting,filice2020crowdsourcing,objectcxr,jaeger2014two}. 
Third, existing large-scale datasets contain a broad range of disease categories in different granularities~\cite{bustos2020padchest,mei2022radimagenet}.
For example, in PadChest~\cite{bustos2020padchest}, disorders exist at different levels, including infiltrates, interstitial patterns, and reticular interstitial patterns. Such variation in granularity poses a great challenge for representation learning. In summary, these datasets fall short of meeting real-world clinical needs, which often involve complex, multimodal, and multi-image data from a single patient. Therefore, constructing a dataset that mirrors the intricacies of actual clinical scenarios is necessary. In comparison, the large-scale radiology dataset we construct covers a large number of categories of diseases, including various common modalities.

The prevailing paradigm in earlier diagnostic models is a specialized model trained on a limited range of disease categories. These models focus on specific imaging modalities and are targeted towards particular anatomical regions. Specifically, ConvNets are widely used in medical image classification due to their outstanding performance, for example, \cite{deepak2019brain, swati2019brain, chowdhury2020can} have demonstrated excellent results on identifying a wide range of diseases. 
Recently, Vision Transformer (ViT) has garnered immense interest in the medical imaging community, numerous innovative approaches ~\cite{dai2021transmed, Wu2023MedKLIPMK, park2021vision, zhang2023knowledge, Wu2023KDiagKD, liu2021automatic, gao2021covid} have emerged, leveraging ViTs as a foundation for further advancements in this field.

The other stream of work is generalist medical foundation models
~\cite{tu2023towards,Wu2023TowardsGF,Zhang2023PMCVQAVI}. 
These models represent a paradigm shift in medical AI, 
aiming to create versatile, comprehensive AI systems capable of handling a wide range of tasks across different medical modalities, by leveraging large-scale, diverse medical data. MedPaLM M~\cite{tu2023towards} reaches performance competitive with or exceeding the state-of-the-art (SOTA) on various medical benchmarks, demonstrating the potential of the generalist foundation model in disease diagnosis and beyond. RadFM~\cite{Wu2023TowardsGF} is the first medical foundation model capable of processing 3D multi-image inputs, demonstrating the versatility and adaptability of generalist models in processing complex imaging data. However, the development of GMAI models requires substantial computational power, which is often impractical for exploring sophisticated algorithms in academic labs.

In contrast, leveraging our proposed RP3D-DiagDS dataset, our final model RadDiag is capable of performing universal disease diagnosis and classification tasks within a constrained parameter scope, transcending the limitations of focusing on specific anatomies or disease categories. This capability offers a more versatile and efficient approach to healthcare management, resulting in significant clinical and research implications compared with precious AI models trained on former diagnostic datasets as discussed in Supplementary Section E. Our impacts and contributions can be summarized in the following three aspects:

% Our dataset RP3D-DiagDS compared with former diagnosis datasets (more details in Supplementary Comparison with other datasets) has great clinical and research values as shown in the following three points.
% \xiaoman{what details?}\qiaoyu{qiaoyu: fixed}

\noindent \textbf{Model Development for Case-Level Diagnosis.} 
The RP3D-DiagDS dataset facilitates the creation of AI models capable of case-level diagnosis utilizing multi-modal and multi-scan inputs. 
This is a critical advancement since patients often undergo multiple radiological examinations across different medical departments during their treatment. 
Traditional disease classification models, which typically process single-image scans, are inadequate for such complex scenarios. Our dataset, with its case-level labels, promotes the development of comprehensive diagnostic AI systems, thereby addressing a significant gap in clinical practice.
% With the case-level labels, our dataset promotes and monitors the progress in developing a case-level diagnosis AI system, enabling comprehensive diagnosis. In clinical practice, patients may get multiple radiologic examinations during their long-term treatment progress from various medical departments. However, existing work on disease classification accepting one image scan, can hardly handle such circumstances, hindering the development of accurate and comprehensive diagnostic models. 
    
\textbf{Long-Tailed Rare Disease Diagnosis.} 
Unlike existing datasets that primarily cover common diseases, the RP3D-DiagDS encompasses over 5,000 disorders, including a significant number of rare diseases. This focus on long-tailed rare disease diagnosis is particularly important for clinical applications, as AI systems can readily assist in diagnosing common diseases but often fall short in identifying rare conditions. 
Our dataset thus plays a crucial role in enhancing the diagnostic process for rare diseases, which are typically more challenging to diagnose due to their lower prevalence.

% With over 5000 disorders, more rare diseases are contained in our label space. Compared with existing diagnosis works that focus on common general disease classes, our efforts on long-tailed rare classes are more vital for clinical usage. For common diseases, AI diagnosis system, usually, can only help accelerate diagnosis procedure for clinicians, while the hint on rare classes is also critical.
    
\noindent \textbf{Value of Online Radiology Resources.} 
The RP3D-DiagDS dataset, collected from the Internet, serves as a valuable resource for training disease diagnosis model, especially for local medical centers dealing with rare diseases. Given the scarcity of cases for such diseases, our dataset's ability to support few-shot learning scenarios is of paramount importance. It demonstrates the value that, such publicly shared medical data on the Internet represents a colossal and high-quality resource, holding the potential to pave the way for constructing generalist models aimed at advancing AI4healthcare. 

%This approach not only enhances the diagnosis and treatment of rare diseases but also signifies a stride towards the democratization of medical knowledge, fostering the development of AI models that are more inclusive and representative of diverse medical conditions.
% This facilitates the development of AI models that can perform effectively even with a limited number of training samples, thereby significantly advancing the diagnosis and treatment of rare diseases.
%\weidi{this paragraph needs to be edited, to demonstrate the message that `publicly shared medical data on the Internet is a tremendous and valuable resource that can potentially support building generalist model towards AI4healthcare'.}\qiaoyu{fixed}

% In clinical application, local medical centres can leverage this superiority even with a few training samples. This is remarkable, especially considering that, for rare diseases, only a very few cases could be accessible.

Despite the effectiveness of our proposed dataset, and architecture, 
there remains certain limitations: 
{\em First}, on model design, in the fusion step, we can use more image tokens to represent a scan rather than a pooled single vector. The latter may lead to excessive loss of image information during fusion. The model size can be further increased to investigate the effect of model capacities; 
{\em Second}, new loss functions should be explored to tackle such large-scale, long-tailed disorder/disease classification tasks. 
{\em Third}, on the disorder to ICD-10-CM mapping progress, the annotators are labeled on class-name level, {\em i.e.}, only disorder names are provided, causing some ambiguous classes unable to find strict corresponding ICD-10-CM codes. 
Though we mark out this class in our shared data files, if providing more case-level information, the mapping could be more accurate. We treat this as future work.

\section{METHODS}

In this section, we will detail the following aspects separately: 
dataset construction, model design, and experiment implementation.

\subsection{Dataset Construction}
In this section, we present the details of our dataset, \textbf{RP3D-DiagDS}, that follows a similar procedure as RP3D~\cite{Wu2023TowardsGF}. Specifically, cases in our dataset are sourced from the Radiopaedia website~\cite{Radiopaedia} -- a growing peer-reviewed educational radiology resource website, that allows clinicians to upload 3D volumes to better reflect real clinical scenarios. Additionally, all privacy issues have already been resolved by the clinicians at uploading time. It is worth noting that, unlike RP3D~\cite{Wu2023TowardsGF} that contains paired free-form text description and radiology scans for visual-language representation learning, here, we focus on multi-modal, multi-anatomy, and multi-label disease diagnosis~(classification) under extremely unbalanced distribution. 

Overall, the proposed dataset contains \change{40,936 cases, 
of 195,010 images} from 9 diverse imaging modalities and 7 human anatomy regions, note that, each case may contain images of multiple scans. The data covers 5,568 different disorders, that have been manually mapped into 930 ICD-10-CM~\cite{ICD10} codes. 

% To explore the accuracy of classification models on the mixed medical radiologic images of various diseases, we establish a large-scale radiologic diagnostic dataset called RP3D-DiagDB, which contains 60k cases covering 5658 diseases on 916 ICD-10-CM\footnote{\url{https://www.ICD-10data.com/ICD-10CM/Codes}} classes, containing xxx images from 9 modalities and 7 anatomies. \xiaoman{list the modality and anatomy here?}
% To the best of our knowledge, RP3D-DiagDB is the first multi-disease, cross-modality, and cross-anatomy radiologic diagnostic dataset.
% \xiaoman{mention the dataset keeps growing or not?}

\begin{figure}[t]
    \centering
    \includegraphics[width=0.99\linewidth]{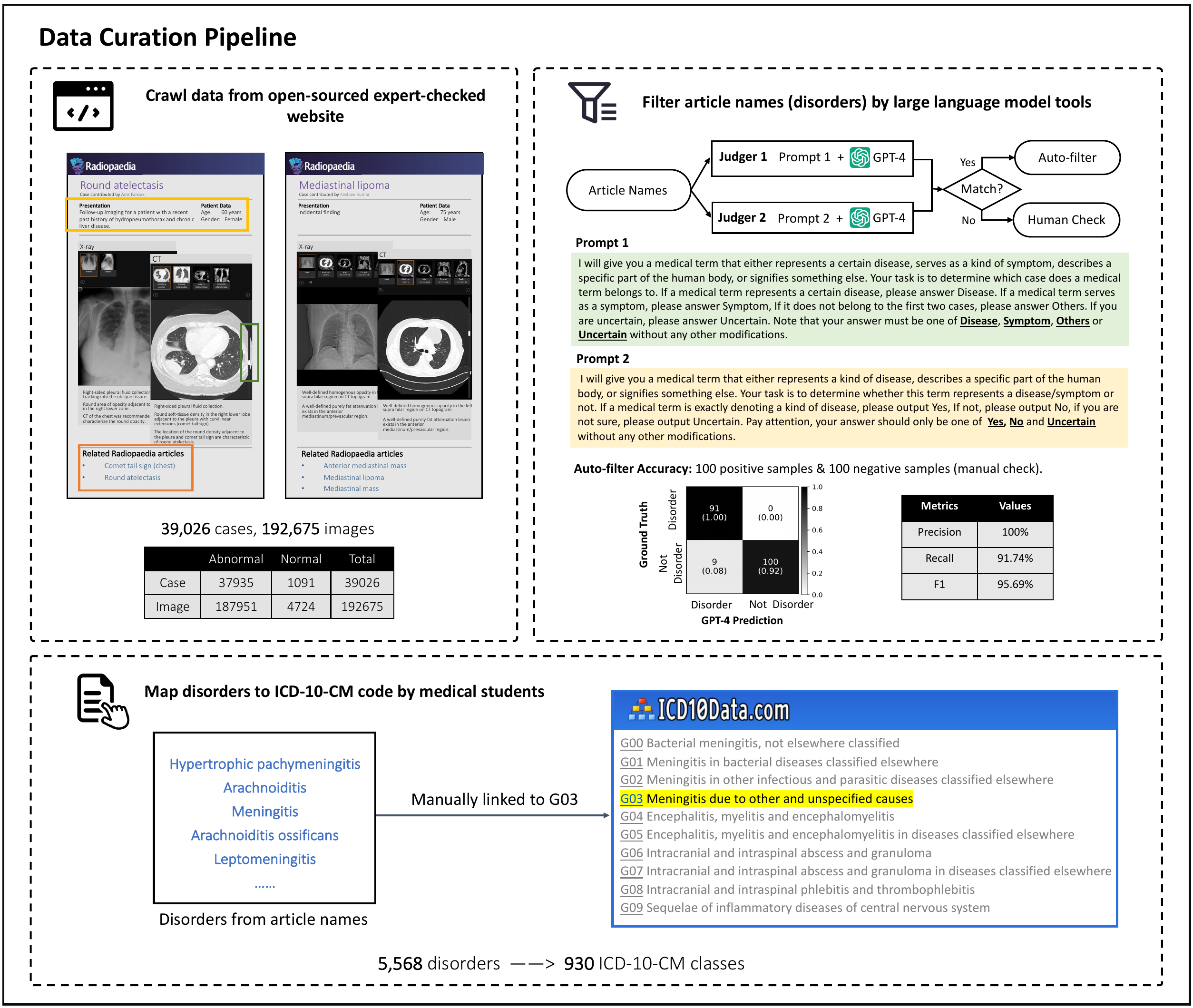}
    \caption{\textbf{The data collection pipeline.} First, we crawl data from an open-sourced and expert-checked website, including case descriptions, 2D/3D radiology images, and related Radiopaedia articles. Then, we filter article names by large language model tools (GPT-4) and check manually if mismatch. Finally, we map the article names into ICD-10-CM codes, cross-checked by a 10-year clinician.}
    \label{fig:process}
\end{figure}

We describe the procedure of our data curation process, shown in Figure~\ref{fig:process}. Specifically, we collect three main components from each case on Radiopaedia webpage, namely, ``Patient data'', ``Radiology images'', and ``Articles'' ~(example webpage shown in Figure~\ref{fig:process}). 
``Patient data'' includes brief information about this patient, for example, age, gender, {\em etc}. ``Radiology images'' denote a series of radiology examination scans. \change{Notably, when uploading the cases, the physicians have also picked a default \textbf{key slice} to support diagnosis for each scan in  ``Radiology images'', we also crawl this as a piece of extra information.} 
Figure~\ref{fig:multi-img}a provides a statistical analysis on the number of images within one single case. ``Articles'' contains links to related articles named with corresponding disorders, which are treated as diagnosis labels and have been meticulously peer-reviewed by experts in Radiopaedia Editorial Board\footnote{\url{https://radiopaedia.org/editors}}.

To start with, we collect 50,970 cases linked with 10,670 articles from the Radiopaedia website. However, naively adopting the article titles as disorders may lead to ambiguities from two aspects: 
(i) not all articles are related to disorders, 
(ii) article titles can be written with different granularity, 
for example, ``pneumonia'' and  ``bacterial pneumonia'' can be referred to as different disorders, though they should ideally be arranged in a hierarchical structure, (iii) normal cases from Radiopaedia are not balanced in modalities and anatomies. Next, we discuss the 3-stage procedure proposed to alleviate the above-mentioned challenges.

\vspace{3pt}
\noindent \textbf{Article Filtering.}
We leverage the GPT-4~\cite{OpenAI2023GPT4TR} to automatically filter the article list, keeping those referring to disorders. Specifically, taking inspiration from the self-consistency prompts~\cite{wang2022self}, we design two different query prompts with similar meanings, as shown in Figure~\ref{fig:process}. An article name is labeled as a disorder if GPT-4 consistently gives positive results from both prompts, while for those GPT-4 gives inconsistent results, we manually check them. To measure the quality of filtering, we randomly sample a portion of data for manual checking to control its quality. The confusion metrics are shown in Figure~\ref{fig:process}. The $100\%$ precision score indicates our auto-filtering strategy, can strictly ensure the left ones to be disordered. 
Eventually, 5342 articles can pass the first auto-criterion and 226 pass the second manually checking, resulting in 5,568 disorder classes, as shown in Figure~\ref{fig:process}b. Cases not linked to any articles are excluded, ultimately yielding 38,858 cases.

% \begin{figure}[t]
%     \centering
%     \includegraphics[width=1\linewidth]{figure/ICD-FLOW.pdf}
%     \caption{The article filtering pipeline. 
%     In (a), we show the two GPT prompts and (b) is the whole pipeline where we leverage the two prompts to query GPT-4 twice and only the consistent results will be used for auto-filtering, otherwise, we will check it manually. (c) shows the accuracy of the auto-filtering cases. Note that, all the 200 article samples evaluated in (c) have a consistent prediction under the two GPT-4 prompts.}
%     \label{fig:double check}
% \end{figure}

\vspace{3pt}
\noindent \textbf{Mapping Disorders to ICD-10-CM.}
Upon getting disorder classes, they may fall into varying hierarchical granularity levels, here, we hope to map them to internationally recognized standards. For this purpose, we utilize the \textbf{International Classification of Diseases, Tenth Revision, Clinical Modification (ICD-10-CM)} codes, which is a customized version of the ICD-10 used for coding diagnoses in the U.S. healthcare system. 
The ICD was originally designed as a healthcare classification system, providing a system of diagnostic codes for classifying diseases, including nuanced classifications of a wide variety of signs, symptoms, abnormal findings, complaints, social circumstances, and external causes of injury or disease. 
Specifically, we hire ten medical Ph.D. students to manually map the article titles into ICD-10-CM codes. After cross-checking by a ten-year clinician, we have mapped the 5,568 disorders into the corresponding ICD-10-CM codes. 
We unify various disorders into the first hierarchical level of ICD-10-CM code tree, \textbf{resulting in 930 ICD-10-CM classes}. For example, both ``J12.0 Adenoviral pneumonia'' and  ``J12 Viral pneumonia'' are mapped to the code ``J12 Viral pneumonia, not elsewhere classified'', dismissing the ambiguities caused by diagnosis granularity. Note that, we will provide the ICD-10-CM codes along with the original 5,568 classes, each accompanied by its corresponding definition, thus, the dataset can be employed for training diagnosis or visual-language models.

\vspace{3pt}
\noindent \textbf{Adding Normal Cases.}  
Despite there exist normal cases on Radiopaedia\footnote{\url{https://radiopaedia.org/articles/normal-imaging-examples}}, covering most anatomies and modalities, \change{we additionally collect more normal cases from additional external datasets, including Brain-Tumor-17, CheXpert, Vertebrase-Xray, KneeMRI, and VinDr-Mammo, and} MISTR\footnote{\url{https://mistr.usask.ca/odin/}}, with images available for research under a CC-BY-NC-SA 4.0 license, \emph{i.e.}, simply adding the normal case part introduced in them. The expanded normal cases include \change{2503} cases.

Finally, \textbf{we get \change{40,936 }cases containing \change{195,010} images labeled by 5,568 disorder classes and 930 ICD-10-CM classes and will continually maintain the dataset, growing the case number.} 

\subsection{Model Design}
\label{modelArch}
Here, we aim to initiate a preliminary investigation on computational architectures for large-scale, long-tailed disease diagnosis on radiology, specifically, we have defined the problem of case-level multi-label classification, and elaborate the architecture in Sec.~\ref{architecture}, and knowledge-enhanced training strategy in Sec.~\ref{knowledge_enhanced}.

\subsubsection{Architecture}
\label{architecture}
\begin{figure}[t]
    \centering
    \includegraphics[width=1\linewidth]{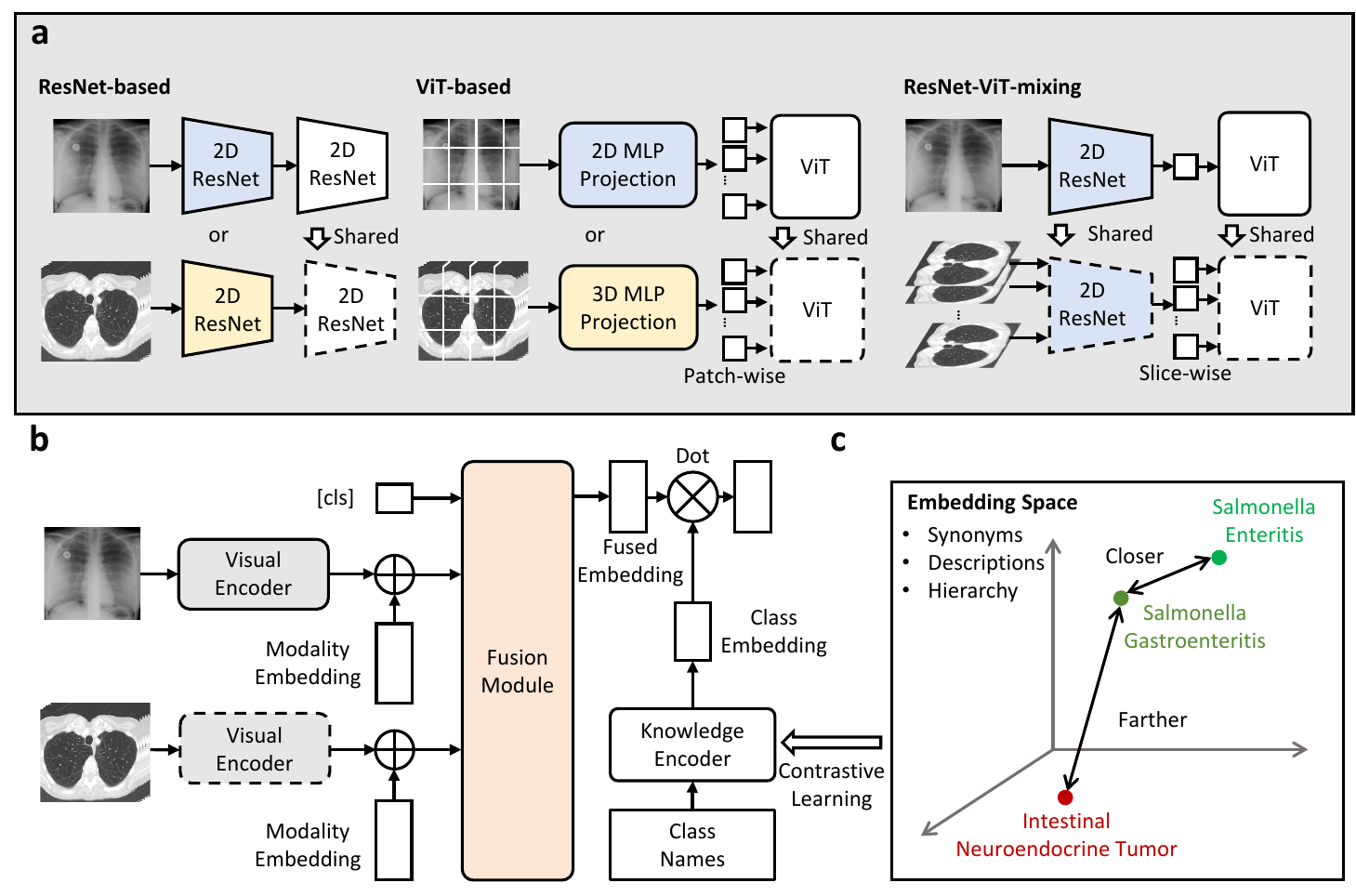}
    \caption{\textbf{The overview of our method.} Three parts demonstrate our proposed visual encoders and fusion module, together with the knowledge enhancement strategy respectively. \textbf{a} The three types of vision encoder, \emph{i.e.}, ResNet-based, ViT-based , and ResNet-ViT-mixing. \textbf{b} The architecture of the fusion module. The figure shows the transformer-based fusion module, enabling case-level information fusion. \textbf{c} The knowledge enhancement strategy. We first pre-train a text encoder with extra medical knowledge with contrastive learning, leveraging synonyms, descriptions, and hierarchy. Then we view the text embedding as a natural classifier to guide the diagnosis classification.
    %\weidi{the order should be reversed, do vision encoder first, then fusion.}
    %\weike{change the font\textasciitilde}
    %\chaoyi{fixed :-)}
    }
    \label{fig:method}
\end{figure}

Our proposed \change{\textbf{RadDiag}} architecture consists of two key components: 
(\textit{i}) a visual encoder that processes 2D or 3D input scan; (\textit{ii}) a transformer-based fusion module, merging all information to perform case-level diagnosis.

% We consider two methods for long-tailed disease diagnosis at case level, \emph{i.e.}, (\textit{i}), a fusion module, (\textit{ii}) knowledge enhanced pipeline. In the former strategy, we propose a transformer-based fusion module and feed all scans from the same case
% as input and . In latter one, we further propose a knowledge encoder and guide the classification network with it.

\noindent \textbf{Visual Encoder.} We consider two popular variants of the visual encoder, namely, ResNet~\cite{he2016identity} and ViT~\cite{dosovitskiy2020image}, as shown in Figure~\ref{fig:method}a. The visual encoding progress can generally be formulated as:
\begin{equation}
    v_i = \Phi_\text{visual}(x_i) \in \mathcal{R}^{d}.
\end{equation}
where $x_i \in \mathcal{R}^{C \times H \times W \times (D)}$ denotes the input scan, $C, H, W$ refer to the image channel, height, and width of the input scan respectively. $D$ is optional, and only available for 3D input scans. The main challenge for architecture design comes from the requirement to process scans in both 2D and 3D formats. 
Here, we train separate \textbf{normalization modules} to convert the 2D or 3D inputs into feature maps of the same resolution and further passed into a \textbf{shared encoding module}. \change{We here adopt three visual encoding variants to handle the challenge:}

%For ResNet-based architecture, we adopt a series of 2D or 3D convolution blocks for normalization, 
%while for ViT, as it enables to handle sequences of variable tokens, the normalization is simply based by a linear patch projection layer.

\vspace{-0.15cm}
\begin{itemize}
\setlength\itemsep{0.15cm}
    \item \textbf{ResNet-based.} For 3D scans, they are first fed into 3D ResNet, followed by average pooling on depth to aggregate the information of the extra dimension; while for 2D scans, they are fed to the 2D ResNet to perform the same down-sampling ratio on height and width as for 3D.  
    After normalization, both 2D and 3D scans are transformed into feature maps with the same resolution, $f_i \in \mathcal{R}^{d_{\text{res}} \times h \times w}$, where $d_{\text{res}}$ is the intermediate feature dimension and $h,w$ denote the normalized size. Then, the feature maps are passed into a \textbf{shared} ResNet to get the final visual embedding.
    
    \item \textbf{ViT-based.} For 3D scans, we convert the input volume into a series of non-overlapped 3D cubes, and pass them into MLP projection layers, to get vector embeddings; while for 2D scans, the input scan is broken into 2D patches, and projected into vector embeddings with another set of MLP layers. As ViT enables handling sequences of variable tokens, we can now pass the resulting vectors into a \textbf{shared} ViT to get the final visual embedding by confusing the feature embedding from arbitrary \change{patch tokens}. For position embedding, we adopt two sets of learnable position embeddings, one for 2D input, and the other for 3D, further indicating the network about the input dimension.
    
    \item \textbf{ResNet-ViT-mixing.} \change{Inspired by former works on pseudo-3D residual networks~\cite{Qiu_2017_ICCV}. We, herein, also design a new 2D + 1D network architecture mixing both ResNet and ViT structures to encode 3D or 2D scans consistently. Specifically, for 3D scans, we treat them as a sequence of 2D slices, and for 2D scans, we also treat them as a sequence of 2D slices of length 1. We first adopt 2D ResNet to get normalized feature embedding per slice as $f_i \in \mathcal{R}^{d_{\text{res}} \times h \times w}$. We then use transformer architecture to perform aggregation on the feature of slice tokens, and thus is similar to ViT-based method. Thanks to the flexibility of transformer architecture, the model can handle arbitrary number of tokens, obtaining the final visual embedding by integrating features from varying slice feature tokens.}
\end{itemize}

\noindent \textbf{Fusion Module.}
For case-level diagnosis, we propose to aggregate information from multiple scans with a trainable module. As shown in Figure~\ref{fig:method}b, we adopt transformer encoders, specifically, we first initialize a set of \textbf{learnable} modality embeddings, denoted as $\{ p_1, \dots, p_M\}$, where $M$ denotes the total number of possible imaging modalities. 
Given certain visual embedding~($v_i$) from modality $j$, we first add the corresponding modality embedding~($p_j$) to it, indicating which radiologic modality it is from. 
Then, we feed all visual embeddings from one case into the fusion module, and output the fused visual embedding $v_\text{fuse} \in \mathcal{R}^d$, from the ``[CLS]'' token, similar to paper~\cite{dosovitskiy2020image}, denoted as:
\change{
\begin{equation}
    v_\text{fuse} = \Phi_\text{fuse}(v_1+p_{m_1}, v_2+p_{m_2}, \cdots, v_S+p_{m_S}) \in \mathcal{R}^d,
\end{equation}
}
\change{where $v_i$ denotes the visual embedding for each image scan and $m_i$ denotes the its corresponding imaging modality, like MRI or CT.} 
Till here, we have computed the case-level visual embedding, which can be passed into a classifier for disease diagnosis. 
We adopt a knowledge-enhanced training strategy~\cite{Wu2023KDiagKD, zhang2023knowledge}, 
that has shown to be superior in long-tailed recognition problems, 
as detailed in the following section.

%Then by adopting an MLP-based classifier, we can get the final prediction $p \in \mathcal{R}^c$. But, the simple MLP classifier will dismiss the information from class names, which can indicate which classes are similar and help long-tailed or few-shot classes more. Thus we further propose the knowledge enhancement strategy.

\subsubsection{Knowledge-enhanced Training}
\label{knowledge_enhanced}

With knowledge enhancement, we hope to leverage the rich medical domain knowledge to enhance the long-tailed classification. 
Our key insight is that the long-tailed diseases may fundamentally have some shared symptoms or radiological pathologies with the head classes, which could be encoded in text format.  In detail, we first pre-train a text encoder with medical knowledge, termed as knowledge encoder, where names of similar disorders or diseases are projected to similar embeddings, for example, ``lung disease'' is closer to ``pneumonia'' in the embedding space than  ``brain disease''. Then, we freeze the text embeddings, and use them to guide the training of the vision encoder, as shown in Figure~\ref{fig:method}c. 

\noindent \textbf{Knowledge Encoder Pre-training.}
We leverage several knowledge bases to pre-train a knowledge encoder, including Radiopaedia, ICD10-CM, and UMLS. Specifically, for each disorder term, we collect its definitions, radiologic features from Radiopaedia articles, synonyms, clinical information, and hierarchy structure from ICD10-CM, as well as definitions from UMLS. We aim to train the text encoder with the three considerations, 
\emph{i.e.}, \textbf{synonyms}, \textbf{hierarchy}, and \textbf{descriptions}. First, if two terms are identified as similar synonyms, we expect them to be close in text embedding space, like ``Salmonella enteritis'' and ``Salmonella gastroenteritis''. Second, in the context of hierarchical relationships, we expect that if a disease is a fine-grained classification of another disease, its embeddings should be closer than those of unrelated diseases. For example, ``J93, Pneumothorax and air leak'' should exhibit closer embeddings with ``J93.0, Spontaneous tension pneumothorax'' than with ``J96.0, Acute respiratory failure''. Third, for terms associated with descriptions, radiological features, or clinical information, we expect their embeddings to be close to the text embedding space.  For example, ``Intestinal Neuroendocrine Tumor'' and ``A well-differentiated, low or intermediate grade tumor with neuroendocrine differentiation that arises from the small or large intestine''.    

We start from an off-the-shelf text encoder, 
namely, MedCPT-Query-Encoder~\cite{jin2023medcpt}, 
and adopt contrastive learning for further finetuning. 
Given a target medical terminology name encoded by the text encoder, denoted as $f_\text{tar}$, its corresponding medical texts, \emph{i.e.}, synonyms, containing terminologies or related descriptions are treated as positive cases. Similarly, we encode them with the text encoder, denoted as $f^{+}$, and other non-related text embedding is treated as negative cases $f^{-}$. \textbf{Note that}, we keep positive and negative cases consistent in format, \emph{e.g.}, when the positive case is related to a description sentence, the negative cases are always some non-related description sentences instead of the short name words. The final objective can be formulated as:
\begin{equation}
    \mathcal{L}_\text{knowledge} = -\log \frac{e^{f_\text{tar}^T \cdot f^{+}/\tau}}{\sum_{n=1}^{N}e^{f_\text{tar}^T \cdot f^{-}_n/\tau}},
\end{equation}
where $\tau$ refers to the temperature, and $N$ denotes total sampled negative cases. By optimizing the contrastive loss we can further finetune the text encoder, resulting in a knowledge encoder, termed as $\Phi_{k}$.

\vspace{3pt}\noindent \textbf{Knowledge-guided Classification.}
After training the knowledge encoder, we use it to encode the disorder/disease names into text embeddings, for example, denoting the names as $\{T_1, T_2, \dots, T_c\}$, where $T_j$ is a disorder/disease name like ``pneumonia'' or ``lung tumor''. 
We embed these free texts with the knowledge encoder as:
\begin{equation}
    t_j = \Phi_k(T_j) \in \mathcal{R}^{d},
\end{equation}
where $t_j$ denotes the text embeddings that are used for computing dot product with visual embeddings for case-level diagnosis:
\begin{equation}
    p = v_\text{fuse} \cdot t \in \mathcal{R}^c,
\end{equation}
where $p$ is the final result.
% \weidi{is it cosine similarity or just dot product ? cosine similarity needs L2 norm.} 
We use classical binary cross entropy~(BCE) loss as the final training objective, denoted as:
\begin{equation}
    \mathcal{L} = - \sum_{i=1}^{c}  \mathcal{Y}_i \cdot \log(p_i) + (1 - \mathcal{Y}_i) \cdot \log(1 - p_i).
\end{equation}

\subsection{\change{Evaluation Details}}
\change{In this section, we will describe our evaluation details. First, we build up an internal test-set based on our proposed \textbf{RP3D-DiagDS}, serving for assessing the models' ability on case-level, multi-modal, multi-scan, long-tailed disease diagnosis. Then we introduce more external datasets for assessing the transferring ability of our final model on other clinical benchmarks. 
Lastly, we introduce the used evaluaiton metrics, along with the compared baselines. }

\subsubsection{The Internal Test-Set}
\change{To start with, for better assessing the model's ability on case-level diagnosis, we split a test set from our own proposed datasets, \textbf{RP3D-DiagDS} to perform internal evaluation. Specifically,}
% \subsubsection{Benchmark Statistics}
we randomly split our dataset into training~(train and validation) and testing sets in a (7:1):2 ratio
%, \change{termed as \textbf{RP3D-DiagDS-Test}}
. 
We treat the class with $[100:]$ positive cases as head classes, $[30,100)$ cases as medium classes, and $[:30)$ cases as tail classes. 
Consequently, from a total of 40,936 cases in our proposed RP3D-DiagDS, the training set comprises 29,536 cases, and the test set includes 7,772 cases. 
Following on the dataset construction procedure, we perform classification tasks on two class sets at different granularities: (i) 5568 disorders + normal; (ii) 930 ICD-10-CM codes + normal. On different sets, we will have different head/medium/tail class sets following our definition. Note that, the normal class is always treated as one of the head classes. 
Next, we will only discuss the abnormal classes.

\vspace{3pt}
\noindent \textbf{Disorders.} 
At the disorder level, there are 85 head classes, 470 medium classes, and 5014 tail classes. Notably, as the case number of some tailed classes is extremely small, some classes may only appear in training or test sets, but not both. This issue only happens to the tail classes, while our knowledge encoder~(as detailed in Sec.~\ref{knowledge_enhanced}) enables embedding for unseen diseases, ensuring that evaluation is not affected by such challenge. As a result, there are 5,260/3,391 classes in our training and testing split, respectively.

\vspace{3pt}
\noindent \textbf{ICD-10-CM.} 
At the ICD-10-CM level, disorders with similar meanings have been mapped to the same codes, for example, ``tuberculosis'', and ``primary pulmonary tuberculosis'' all correspond to ``A15 respiratory tuberculosis''. This merge operation results in more cases in one class and, consequently more head classes. Specifically, we have 165 head classes, 230 medium classes, and 536 tail classes. Similarly, some tail classes may only be in training or testing split, resulting in 894/754 classes for our final training and testing splits, respectively. 

It is important to point out that while we ensure the inclusion of head and medium classes in both the training and test sets, some tail classes are inevitably exclusive from certain divisions, which is treated as our future work to collect more samples of such disorders. We report the results separately for head/medium/tail classes while focusing primarily on the head classes. 

% In addition to \textbf{RP3D-DiagDS-Test}, we also propose a relatively small subset by only keeping the top 200 disorder classes with most cases, namely \textbf{RP3D-DiagDS-Test}. 
% compare various configurations more quickly, we conduct ablation studies on a subset of original dataset, comprising 200 disorder categories with most cases, termed as SubSet@200. 

\subsubsection{The External Test-Sets}
In addition to evaluating on our own testset, 
we also consider zero-shot or finetuning evaluation on other external datasets, demonstrating its transferring abilities on image distribution shift and label space change. In the following sections, we start by introducing the used external datasets, that cover various medical imaging modalities and anatomies. Then we detail the fine-tuning settings. We choose an external evaluation dataset with the following principles:
\vspace{-0.15cm}
\begin{itemize}
\setlength\itemsep{0.15cm}
    \item \textbf{Imaging Modalities.} 
    We hope to cover most radiologic modalities in our external evaluation, \emph{e.g.}, 2D X-ray, 2D CT/MRI slices, and 3D CT/MRI scans demonstrating our model can benefit all of them;
    
    \item \textbf{Human Anatomies.} 
    We hope to cover many human anatomies, \emph{e.g.}, brain, head and neck, chest, spine abdomen, and limb, demonstrating our model can benefit all of them;
    
    \item \textbf{Label Space.} 
    We hope to cover two cases, \emph{i.e.}, seen and unseen classes. Seen classes refer to those labels that have appeared in RP3D-DiagDS. Conversely, unseen extra classes denote those not included.
\end{itemize}

As a result, we pick the following datasets and report AUC scores to compare with others on the external evaluation.
\vspace{-0.15cm}
\begin{itemize}
\setlength\itemsep{0.15cm}
\item \textbf{CXR14~\cite{wang2017chestx}} is a widely-used chest X-ray~(2D) diagnosis dataset containing 112,120 frontal-view X-ray images of 30,805 (collected from the year 1992 to 2015) unique patients with 14 finding labels. We follow its \textbf{official split} and evaluate the SOTA~\cite{Wu2023MedKLIPMK} on the split.

\item  \textbf{VinDr-Spine~\cite{vindrmammo}} is a spine X-ray~(2D) diagnosis dataset comprising 10,469 images from 5,000 studies. We follow K-Diag~\cite{Wu2023KDiagKD} to use the 8 unique finding labels and the \textbf{official split}. 

\item \textbf{VinDr-Mammo~\cite{vindrmammo}} is a mammography~(2D) diagnosis dataset comprising 20,000 images (5,000 four-view scans). Each scan was manually annotated with a 5-level BI-RADS score. We view this as a multi-class classification task with the \textbf{official split}.

\item \textbf{ADNI~\cite{jack2008alzheimer}} is a 3D brain MRI dataset focused on Alzheimer's disease comprising 112141 images, including AD, MCI, CN and other classifications. We \textbf{randomly split} it followed 63,846/15,962 for training and testing and reproduced the state-of-the-art method on it.

\item \textbf{MosMedData~\cite{morozov2020mosmeddata}} is a 3D chest CT dataset on 5-level COVID-19 grading comprising 1,110 images. We follow the \textbf{official split}. We randomly split it following 888/222 for training and testing and reproduced the state-of-the-art method on it.

\item 
\textbf{NSCLC (Radiogenomics)~\cite{bakr2018radiogenomic}} is a 3D chest dataset containing CT and PET scans. There are a total of 211 subjects with Non-small Cell Lung Cancer. We split the dataset based on the overall cancer stage (4 classes). \change{The training and testing set are \textbf{randomly split} as 169 and 42.}

\item 
\textbf{TCGA (Merged)} is a 3D chest dataset composed of three chest-related data sets: TCGA-LUSC~\cite{tcgalusc}, TCGA-LUAD~\cite{tcgaluad} and TCGA-BRCA~\cite{tcgabrca}. It contains three different cancers and a total of 356 cases, \change{which is \textbf{randomly split} into 285 for training and 71 for testing.}

\item 
\textbf{ISPY1~\cite{hylton2016neoadjuvant}} is a breast 3D MRI dataset from patients in the I-SPY 1/ACRIN 6657 trials. It contains a total of 222 subjects with Breast Cancer, and we selected 755 samples among them as the data set. \change{We \textbf{randomly split} the 604 for training and 151 for training.}

\item \change{\textbf{ChestX-Det10~\cite{liu2020chestxdet10}} is a 2D chest X-ray subset of NIH ChestX-14. It contains 3543 images, divided into 10 categories in total. We
follow the \textbf{official split}.}

\item \change{\textbf{CheXpert~\cite{irvin2019chexpert}} is a large dataset of chest X-rays and competition for automated chest x-ray interpretation, which features uncertainty labels and radiologist-labeled reference standard evaluation sets. It contains 224,316 chest radiographs if 65,240 patients, divided into 14 classes. We follow the \textbf{official split}.}

\item 
\change{\textbf{COVID-19-Radio~\cite{chowdhury2020can, rahman2021exploring}} contains 3616 COVID-19 positive cases along with 10,192 Normal, 6012 Lung Opacity (Non-COVID lung infection), and 1345 Viral Pneumonia 2D X-ray images and corresponding lung masks. We \textbf{randomly split} it to 2892 and 1220 for training and testing.}

\item \change{\textbf{IU-Xray~\cite{pavlopoulos2019survey}} is a 2D chest X-ray datasets containing 7430 images in total. It is \textbf{randomly split} into 6674 for training and 756 for testing.}

\item \change{\textbf{LNDb~\cite{Pedrosa2019LNDbAL} }is developed as an external 3D dataset complimentary to LIDCIDRI. It contains 294 CT scans collected retrospectively at the Centro Hospitalar e Universitário de São João (CHUSJ) in Porto. We \textbf{randomly split} it into 235 for training and 59 for testing.}

\item \change{\textbf{PadChest~\cite{bustos2020padchest} }includes more than 160,000 2D chest X-ray images obtained from 67,000 patients that were interpreted and reported by radiologists. The reports were labeled with 174 different radiographic findings, 19 differential diagnoses and 104 anatomic locations organized as a hierarchical taxonomy and mapped onto standard Unified Medical Language System (UMLS) terminology. We use a subset of this dataset and \textbf{randomly split} it into 31242 for training and 7811 for testing.}

\item \change{\textbf{CC-CCII~\cite{zhang2020clinically}} contains 3777 3D volumns. All CT images are classified into novel coronavirus pneumonia (NCP) due to SARS-CoV-2 virus infection, common pneumonia and normal controls. We \textbf{randomly split} the dataset into 3021 for training and 756 for testing.}

\item \change{\textbf{RadChest~\cite{draelos2021machine}} dataset is a large 3D medical imaging dataset developed by Duke University. The full dataset includes 35,747 chest CT scans from 19,661 adult patients. However, only 3,630 chest CT scans are available. There are 83 different labels in it. We \textbf{randomly split} it into 2904 for training and 726 for testing.}

\item \change{\textbf{Brain-Tumor~\cite{msoud_nickparvar_2021}} is a 2D dataset contains 7023 images of human brain MRI images which are classified into 4 classes: glioma, meningioma, no tumor and pituitary. We \textbf{randomly split} it into 5712 for training and 1311 for testing}

\item \change{\textbf{Brain-Tumor-17~\cite{FERNANDO_2023}} has 4449 real 2D MRI images of magnetic resonance imaging of the skull, in axial planes, weighted in T1, T1 with contrast and T2. It is classified into 6 classes: glioma, meningioma, neurocytoma, normal, other types of injuries and schwannoma. We \textbf{randomly split} the dataset into 3560 for training and 889 for testing.}

\item \change{\textbf{POCUS~\cite{born2020pocovid}} is an chest ultrasound dataset containing more than 200 LUS videos labelled with a diagnostic outcome. It can be classified into 3 classes: COVID-19, pneumonia and regular. We treat this dataset as 2D images combination and \textbf{randomly split} into 2230 for training and 558 for testing.}

\item \change{\textbf{MURA~\cite{rajpurkar2017mura}} is a 2D X-ray dataset containing 14,863 musculoskeletal studies of the upper extremity, where
each study contains one or more views and is manually labeled by radiologists as either normal or
abnormal. We \textbf{randomly split} this dataset int 36806 for training and 3197 for testing.}

\item \change{\textbf{KneeMRI~\cite{vstajduhar2017semi}} is a 2D knee MRI dataset gathered retrospectively from exam records made on a Siemens Avanto 1.5T MR scanner. It contains 917 images, which can be classified into 3 classes. We \textbf{randomly split} this dataset into 733 for training and 184 for testing.}

\item \change{\textbf{CT-Kidney~\cite{islam2022vision}} is a 2D CT datasaet collected from PACS from different hospitals in Dhaka, Bangladesh where patients were already diagnosed with having a kidney tumor, cyst, normal or stone findings. It contains 12,446 unique data within it in which the cyst contains 3,709, normal 5,077, stone 1,377, and tumor 2,283. The train set and test set are \textbf{randomly split} as 9973/2473.}

\end{itemize}
% \chaoyi{need to update to the latest dataset list.}\qiaoyu{Done.} \chaoyi{need how we split the dataset}\qiaoyu{Fixed.}

\subsubsection{Baselines For Comparison}
\label{baselines}

\change{We show the comparison between our model and the existing SOTAs on the external public benchmark. In detail, under the \textbf{fine-tuning setting}, we compare with various specialized state-of-the-arts~(SOTAs) for each datasets. 
Due to the large number of external datasets, we herein do not provide the details of the specific designs per method, while directly indicate the corresponding method by citation in the table, and more details of the designs can be found in the corresponding papers. }

\change{In addition to the comparison with various specialized SOTAs, we also compare with other diagnosis foundation models in the \textbf{zero-shot setting}. 
In detail, we pick out related classes from our class set based on the target disease. It is possible that multiple similar classes in our dataset correspond to a single class in the external dataset due to the definition of class granularity, for example ``Tumor'' class in external dataset may be linked with ``Brain Tumor'', ``Chest Tumor'' and so on in our class list. We treat the final prediction as positive if any class in the related class list is predicted as positive. \textbf{Notably}, as such union merging is carried out after thresholding the prediction scores, AUC score can not be calculated here.
For comparison, we consider two types of baseline models:}
\begin{itemize}
\setlength\itemsep{0.15cm}
    \item \change{\textbf{Classification Models.} By training on a wide range of diagnosis data across various diseases, it enables to build a foundation model by simply optimizing the classification loss. Our work falls in this type, and another related work is RadIN~\cite{mei2022radimagenet}. 
    Similarly to our zero-shot adaptation pipeline, we also adopt \textbf{RadIN} to other datasets zero-shotly, by choosing a most related class from its original class list.}
    
    \item \change{\textbf{Contrastive-learning Models. } In addition to using classification labels, leveraging vision-language pairs, also enables to build up a foundation model through optimizing contrastive loss~\cite{chen2020simple}. Inspired by former works~\cite{tiu2022expert,lin2023pmc}, by treating the target disease class name as free-text, these models can also be adapted to zero-shot diagnosis leveraging the cross-modality similarity. We adopt the latest model, \textbf{BiomedCLIP}~\cite{zhang2023biomedclip} form comparison. It is trained with large-scale image datasets sourced from PubMedCentral papers, and can handle various imaging modalities. However, it is worth emphasizing that both \textbf{RadIN} and \textbf{BiomedCLIP} can only handle 2D input. }
\end{itemize}

\subsubsection{Metrics}
In this part, we will describe the evaluation metrics in detail. 
\textbf{Note that}, the following metrics can all be calculated per class. 
For multi-class cases, we all use macro-average on classes to report the scores by default. For example, ``AUC'' for multi-class classification denotes the ``Macro-averaged AUC on classes''. \change{\textbf{We report all metrics in percent~(\%) in tables.}} 

%For simplicity, in the following, we will simply 
%and next, we will introduce each metric just in a certain class case.  

\noindent \textbf{AUC.} Area Under Curve~\cite{bradley1997use} denotes the area under ROC~(receiver operating characteristic) curve. 
This has been widely used in medical diagnosis, due to its clinical meanings and robustness in unbalanced distribution.

\noindent \textbf{AP.} 
Average Precision~(AP) is calculated as the weighted mean of precisions at each threshold. Specifically, for each class, we rank all samples according to the prediction score, then shift the threshold to a series of precision-recall points and draw the precision-recall~(PR) curve, AP equals the area under the curve. 
This score measures whether the unhealthy samples are ranked higher than healthy ones. 
We report Mean Average Precision~(mAP), which is the average of AP of each class.  %\chaoyi{this seems to be dummy with the begginning of this section?}

%Average Precision (AP) refers to the average precision score obtained at various recall levels, achieved by adjusting the classification threshold.
%of all samples across different classes.
% view all the samples on a certain class as a probable list and calculate an AP score.    
%\weidi{we are not gonna include sample-mAP, right ? fix it.}\chaoyi{fixed}

\noindent \textbf{F1, MCC and ACC.} 
F1 score is the harmonic mean of the precision and recall. 
It is widely used in diagnosis tasks. 
ACC denotes the accuracy score.
MCC~\cite{chicco2020advantages} denotes the  Matthews Correlation Coefficient metric. It ranges from $-1$ to $1$ and can be calculated as follow:
\begin{equation*}
    \text{MCC} = \frac{\text{TN} \times \text{TP} - \text{FN} \times \text{FP}}{\sqrt{(\text{TP} + \text{FP})(\text{TP}+\text{FN})(\text{TN}+\text{FP})(\text{TN}+\text{FN})}}
\end{equation*}
All the three metrics need a specific threshold to compute, 
and we choose one by maximizing F1 on the validation set following former papers~\cite{nguyen2022vindr,Wu2023KDiagKD,Wu2023MedKLIPMK, zhang2023knowledge}.

%\weidi{I don't understand the last sentence, what do you mean by `maximizing F1 on the evaluation set.' meaning we are picking best checkpoint on test set ?} \chaoyi{fixed}

\noindent \textbf{Recall@FPR.} 
%better demonstrating the clinical effectiveness of different models, 
We also report the recall scores at different false positive rate~(FPR), \emph{i.e.}, sample points from class-wise ROC curves. Specifically, We report the \textbf{Recall@0.01}, \textbf{Recall@0.05}, \textbf{Recall@0.1} score, denoting the recall scores at 0.01, 0.05, 0.1 FPR respectively.

% \subsubsection{\change{Human Comparison}}

% \chaoyi{write some details like criteria and pipeline. blabla}

\subsection{\change{Implementation}}
\change{In this section, we will introduce implementation details such as the input and output during training and testing, model hyperparameters, and resource consumption information.}

\change{As input to the model, the images are processed into shape of (H, W, D), where H and W represent the image height and width, both set to 256, and D represents the image depth, set to 32. To standardize the processing of both 2D and 3D images, we treat the 3D images as a stack of 2D slices. For images that do not conform to the specified dimensions, we employ bilinear interpolation for the height and width dimensions and uniform resampling for the depth dimension. Finally, we apply feature scaling for normalization. }

\change{Based on the results of the ablation study for our model architecture, we select ResNet-50 and a transformer encoder for the visual encoder component. The features output by ResNet-50 have a shape of (2048, 8, 8). After applying average pooling, these features are passed through a transformer encoder to obtain features with a shape of (33, 2048). We choose the features of the first [cls] token for subsequent fusion among different scans. 
In the transformer encoder used for fusion, the number of heads is set to 4, the number of layers is set to 2, and the output feature dimension is 2048. 
Finally, through interaction with the knowledge embedding and a fully connected layer, we obtain the final prediction results with a shape of class numbers. 
For more detailed information, please refer to our Github repository.}

\change{The training and testing for this work are conducted using  Nvidia A100s~(80G). During training, we apply image-level augmentation transformations. Additionally, considering that for each scan we have the manually picked key slice, we will also randomly replace the 3D scan with its 2D key slice. When both early fusion and data augmentation are activated, the GPU memory usage for training is approximately 46G with a batch size of 1. During inference, no data augmentation or random mix-up strategies are activated. Under this setting, GPU memory usage is approximately 7G with early fusion and a batch size of 1.}
\section{CONCLUSION}
In this work, we mainly focus on the diagnosis of general diseases~(rare diseases included) based on radiology images, which can be formulated as a problem of multi-modal, multi-label long-tailed case-level diagnosis task. 
Specifically, we exploit the online data resources, constructing a new large-scale diagnosis dataset, namely, RP3D-DiagDS, with 40,936 cases~(195,010 scans) labeled with detailed disorders covering 930 ICD-10-CM.  
On model design, we propose one unified architecture that supports both 2D and 3D input, together with a fusion module, to integrate information from multiple scans of one patient. Additionally, we adopt knowledge enhancement training strategy, leveraging the rich medical domain knowledge to improve the radiological diagnosis performance. Our final model demonstrates strong performance on large-scale radiological disease diagnosis, and shows strong transferring ability to various external datasets regardless of their imaging modalities, shooting anatomies, and target classes. 
As a result, we show that publicly shared medical data on the Internet is a tremendous and valuable resource that can potentially support building a generalist model towards AI4healthcare.

\section{DATA AVAILABILITY}
Due to the Radiopaedia license~\footnote{https://radiopaedia.org/licence?lang=us}, 
we do not directly provide the original dataset. 
To obtain the data, readers must first get the permission from Radiopaedia's founders,
then we will share the download link. 
We also provide the model checkpoint in \url{https://huggingface.co/QiaoyuZheng/RP3D-DiagModel} following the apache-2.0 open-source license and all related result numbers can be reproduced based on the provided model checkpoint.

\section{CODE AVAILABILITY}

% Our dataset RP3D-DiagDS can be found in \url{https://huggingface.co/datasets/QiaoyuZheng/RP3D-DiagDS}, 
Our codes can be found in \url{https://github.com/qiaoyu-zheng/RP3D-Diag} under the MIT License, with all training, evaluation, fine-tuning, and zero-shot evaluation information.

\clearpage
%\setlength\bibitemsep{3pt}
%\printbibliography
%\balance
\bibliographystyle{sn-mathphys} % We choose the "plain" reference style
\bibliography{references} % Entries are in the refs.bib file

\begin{thebibliography}{10}\itemsep=-1pt

\bibitem{ICD10}
{ICD10}.
\newblock \url{https://www.icd10data.com/ICD10CM/Codes}.

\bibitem{SHREYA_2023}
Kaggle: Brain mri scans for brain tumor classification.
\newblock \url{https://www.kaggle.com/datasets/shreyag1103/brain-mri-scans-for-brain-tumor-classification}.

\bibitem{msoud_nickparvar_2021}
Kaggle: Brain tumor mri dataset.
\newblock \url{https://www.kaggle.com/dsv/2645886}.

\bibitem{FERNANDO_2023}
Kaggle: Brain tumor mri images 17 classes.
\newblock \url{https://www.kaggle.com/datasets/fernando2rad/brain-tumor-mri-images-17-classes}.

\bibitem{Radiopaedia}
Radiopaedia.
\newblock \url{https://radiopaedia.org}.

\bibitem{al2020dataset}
Walid Al-Dhabyani, Mohammed Gomaa, Hussien Khaled, and Aly Fahmy.
\newblock Dataset of breast ultrasound images.
\newblock {\em Data In Brief}, 28:104863, 2020.

\bibitem{tcgaluad}
B Albertina, M Watson, C Holback, R Jarosz, S Kirk, Y Lee, K Rieger-Christ, and J Lemmerman.
\newblock The cancer genome atlas lung adenocarcinoma collection (tcga-luad)(version 4)[data set].
\newblock {\em The Cancer Imaging Archive}, 2016.

\bibitem{armato2011lung}
Samuel~G Armato~III, Geoffrey McLennan, Luc Bidaut, Michael~F McNitt-Gray, Charles~R Meyer, Anthony~P Reeves, Binsheng Zhao, Denise~R Aberle, Claudia~I Henschke, Eric~A Hoffman, et~al.
\newblock The lung image database consortium (lidc) and image database resource initiative (idri): a completed reference database of lung nodules on ct scans.
\newblock {\em Medical Physics}, 38(2):915--931, 2011.

\bibitem{bakr2018radiogenomic}
Shaimaa Bakr, Olivier Gevaert, Sebastian Echegaray, Kelsey Ayers, Mu Zhou, Majid Shafiq, Hong Zheng, Jalen~Anthony Benson, Weiruo Zhang, Ann~NC Leung, et~al.
\newblock A radiogenomic dataset of non-small cell lung cancer.
\newblock {\em Scientific data}, 5(1):1--9, 2018.

\bibitem{bhat2023aucreshaping}
Sheethal Bhat, Awais Mansoor, Bogdan Georgescu, Adarsh~B Panambur, Florin~C Ghesu, Saahil Islam, Kai Packh{\"a}user, Dalia Rodr{\'\i}guez-Salas, Sasa Grbic, and Andreas Maier.
\newblock Aucreshaping: improved sensitivity at high-specificity.
\newblock {\em Scientific Reports}, 13(1):21097, 2023.

\bibitem{bien2018deep}
Nicholas Bien, Pranav Rajpurkar, Robyn~L Ball, Jeremy Irvin, Allison Park, Erik Jones, Michael Bereket, Bhavik~N Patel, Kristen~W Yeom, Katie Shpanskaya, et~al.
\newblock Deep-learning-assisted diagnosis for knee magnetic resonance imaging: Development and retrospective validation of mrnet.
\newblock {\em PLoS Medicine}, 15(11), 2018.

\bibitem{born2020pocovid}
Jannis Born, Gabriel Br{\"a}ndle, Manuel Cossio, Marion Disdier, Julie Goulet, J{\'e}r{\'e}mie Roulin, and Nina Wiedemann.
\newblock {POCOVID-Net: automatic detection of COVID-19 from a new lung ultrasound imaging dataset (POCUS)}.
\newblock {\em arXiv preprint arXiv:2004.12084}, 2020.

\bibitem{born2021accelerating}
Jannis Born, Nina Wiedemann, Manuel Cossio, Charlotte Buhre, Gabriel Brändle, Konstantin Leidermann, Avinash Aujayeb, Michael Moor, Bastian Rieck, and Karsten Borgwardt.
\newblock Accelerating detection of lung pathologies with explainable ultrasound image analysis.
\newblock {\em Applied Sciences}, 11(2):672, Jan 2021.

\bibitem{bradley1997use}
Andrew~P Bradley.
\newblock The use of the area under the roc curve in the evaluation of machine learning algorithms.
\newblock {\em Pattern Recognition}, 30(7):1145--1159, 1997.

\bibitem{bustos2020padchest}
Aurelia Bustos, Antonio Pertusa, Jose-Maria Salinas, and Maria de~la Iglesia-Vay{\'a}.
\newblock Padchest: A large chest x-ray image dataset with multi-label annotated reports.
\newblock {\em Medical Image Analysis}, 66:101797, 2020.

\bibitem{cai2019breast}
Hongmin Cai, Qinjian Huang, Wentao Rong, Yan Song, Jiao Li, Jinhua Wang, Jiazhou Chen, Li Li, et~al.
\newblock Breast microcalcification diagnosis using deep convolutional neural network from digital mammograms.
\newblock {\em Computational and Mathematical Methods in Medicine}, 2019, 2019.

\bibitem{cardoso2022monai}
M~Jorge Cardoso, Wenqi Li, Richard Brown, Nic Ma, Eric Kerfoot, Yiheng Wang, Benjamin Murrey, Andriy Myronenko, Can Zhao, Dong Yang, et~al.
\newblock Monai: An open-source framework for deep learning in healthcare.
\newblock {\em arXiv preprint arXiv:2211.02701}, 2022.

\bibitem{chen2020simple}
Ting Chen, Simon Kornblith, Mohammad Norouzi, and Geoffrey Hinton.
\newblock A simple framework for contrastive learning of visual representations.
\newblock In {\em International conference on machine learning}, pages 1597--1607. PMLR, 2020.

\bibitem{cheng2015enhanced}
Jun Cheng, Wei Huang, Shuangliang Cao, Ru Yang, Wei Yang, Zhaoqiang Yun, Zhijian Wang, and Qianjin Feng.
\newblock Enhanced performance of brain tumor classification via tumor region augmentation and partition.
\newblock {\em PloS One}, 10(10):e0140381, 2015.

\bibitem{chicco2020advantages}
Davide Chicco and Giuseppe Jurman.
\newblock The advantages of the matthews correlation coefficient (mcc) over f1 score and accuracy in binary classification evaluation.
\newblock {\em BMC Genomics}, 21(1):1--13, 2020.

\bibitem{chowdhury2020can}
Muhammad~EH Chowdhury, Tawsifur Rahman, Amith Khandakar, Rashid Mazhar, Muhammad~Abdul Kadir, Zaid~Bin Mahbub, Khandakar~Reajul Islam, Muhammad~Salman Khan, Atif Iqbal, Nasser Al~Emadi, et~al.
\newblock Can ai help in screening viral and covid-19 pneumonia?
\newblock {\em IEEE Access}, 8:132665--132676, 2020.

\bibitem{dai2021transmed}
Yin Dai, Yifan Gao, and Fayu Liu.
\newblock Transmed: Transformers advance multi-modal medical image classification.
\newblock {\em Diagnostics}, 11(8):1384, 2021.

\bibitem{deepak2019brain}
S Deepak and PM Ameer.
\newblock Brain tumor classification using deep cnn features via transfer learning.
\newblock {\em Computers in Biology and Medicine}, 111:103345, 2019.

\bibitem{dosovitskiy2020image}
Alexey Dosovitskiy, Lucas Beyer, Alexander Kolesnikov, Dirk Weissenborn, Xiaohua Zhai, Thomas Unterthiner, Mostafa Dehghani, Matthias Minderer, Georg Heigold, Sylvain Gelly, et~al.
\newblock An image is worth 16x16 words: Transformers for image recognition at scale.
\newblock {\em arXiv preprint arXiv:2010.11929}, 2020.

\bibitem{draelos2021machine}
Rachel~Lea Draelos, David Dov, Maciej~A Mazurowski, Joseph~Y Lo, Ricardo Henao, Geoffrey~D Rubin, and Lawrence Carin.
\newblock Machine-learning-based multiple abnormality prediction with large-scale chest computed tomography volumes.
\newblock {\em Medical image analysis}, 67:101857, 2021.

\bibitem{filice2020crowdsourcing}
Ross~W Filice, Anouk Stein, Carol~C Wu, Veronica~A Arteaga, Stephen Borstelmann, Ramya Gaddikeri, et~al.
\newblock Crowdsourcing pneumothorax annotations using machine learning annotations on the nih chest x-ray dataset.
\newblock {\em Journal of Digital Imaging}, 33:490--496, 2020.

\bibitem{fraiwan2022using}
Mohammad Fraiwan, Ziad Audat, Luay Fraiwan, and Tarek Manasreh.
\newblock Using deep transfer learning to detect scoliosis and spondylolisthesis from x-ray images.
\newblock {\em Plos one}, 17(5):e0267851, 2022.

\bibitem{gao2021covid}
Xiaohong Gao, Yu Qian, and Alice Gao.
\newblock {COVID-VIT: Classification of COVID-19 from CT chest images based on vision transformer models}.
\newblock {\em arXiv preprint arXiv:2107.01682}, 2021.

\bibitem{he2016identity}
Kaiming He, X. Zhang, Shaoqing Ren, and Jian Sun.
\newblock Deep residual learning for image recognition.
\newblock {\em 2016 IEEE Conference on Computer Vision and Pattern Recognition (CVPR)}, pages 770--778, 2015.

\bibitem{objectcxr}
JF Healthcare.
\newblock {Object-CXR - Automatic detection of foreign objects on chest X-rays}.
\newblock 2020.

\bibitem{DDSM}
{Heath, Michael and Bowyer, Kevin and Kopans, Daniel, and Moore, Richard and Kegelmeyer, W. Philip}.
\newblock The digital database for screening mammography.
\newblock In {\em Proceedings of the Fifth International Workshop on Digital Mammography}, pages 212--218. Medical Physics Publishing, 2001.

\bibitem{data5010014}
Murtadha~D. Hssayeni, Muayad~S. Croock, Aymen~D. Salman, Hassan~Falah Al-khafaji, Zakaria~A. Yahya, and Behnaz Ghoraani.
\newblock Intracranial hemorrhage segmentation using a deep convolutional model.
\newblock {\em Data}, 5(1), 2020.

\bibitem{hylton2016neoadjuvant}
Nola~M Hylton, Constantine~A Gatsonis, Mark~A Rosen, Constance~D Lehman, David~C Newitt, Savannah~C Partridge, Wanda~K Bernreuter, Etta~D Pisano, Elizabeth~A Morris, Paul~T Weatherall, et~al.
\newblock Neoadjuvant chemotherapy for breast cancer: functional tumor volume by mr imaging predicts recurrence-free survival—results from the acrin 6657/calgb 150007 i-spy 1 trial.
\newblock {\em Radiology}, 279(1):44--55, 2016.

\bibitem{irvin2019chexpert}
Jeremy Irvin, Pranav Rajpurkar, Michael Ko, Yifan Yu, Silviana Ciurea-Ilcus, Chris Chute, Henrik Marklund, Behzad Haghgoo, Robyn Ball, Katie Shpanskaya, et~al.
\newblock {CheXpert: A Large Chest Radiograph Dataset with Uncertainty Labels and Expert Comparison}.
\newblock In {\em Proceedings of the AAAI Conference on Artificial Intelligence}, number~01, pages 590--597, 2019.

\bibitem{islam2022vision}
Md~Nazmul Islam, Mehedi Hasan, Md~Kabir Hossain, Md~Golam~Rabiul Alam, Md~Zia Uddin, and Ahmet Soylu.
\newblock Vision transformer and explainable transfer learning models for auto detection of kidney cyst, stone and tumor from ct-radiography.
\newblock {\em Scientific Reports}, 12(1):1--14, 2022.

\bibitem{jack2008alzheimer}
Clifford~R Jack~Jr, Matt~A Bernstein, Nick~C Fox, Paul Thompson, Gene Alexander, Danielle Harvey, Bret Borowski, Paula~J Britson, Jennifer L.~Whitwell, Chadwick Ward, et~al.
\newblock The alzheimer's disease neuroimaging initiative (adni): Mri methods.
\newblock {\em Journal of Magnetic Resonance Imaging: An Official Journal of the International Society for Magnetic Resonance in Medicine}, 27(4):685--691, 2008.

\bibitem{jaeger2014two}
Stefan Jaeger, Sema Candemir, Sameer Antani, Y{\`\i}-Xi{\'a}ng~J W{\'a}ng, Pu-Xuan Lu, and George Thoma.
\newblock Two public chest x-ray datasets for computer-aided screening of pulmonary diseases.
\newblock {\em Quantitative Imaging in Medicine and Surgery}, 4(6):475, 2014.

\bibitem{jin2023medcpt}
Qiao Jin, Won Kim, Qingyu Chen, Donald~C Comeau, Lana Yeganova, W~John Wilbur, and Zhiyong Lu.
\newblock Medcpt: Contrastive pre-trained transformers with large-scale pubmed search logs for zero-shot biomedical information retrieval.
\newblock {\em Bioinformatics}, 39(11):btad651, 2023.

\bibitem{johnson2019mimic}
Alistair~EW Johnson, Tom~J Pollard, Seth~J Berkowitz, Nathaniel~R Greenbaum, Matthew~P Lungren, Chih-ying Deng, Roger~G Mark, and Steven Horng.
\newblock Mimic-cxr, a de-identified publicly available database of chest radiographs with free-text reports.
\newblock {\em Scientific Data}, 6(1):317, 2019.

\bibitem{joyce2023explainable}
Dan~W Joyce, Andrey Kormilitzin, Katharine~A Smith, and Andrea Cipriani.
\newblock Explainable artificial intelligence for mental health through transparency and interpretability for understandability.
\newblock {\em npj Digital Medicine}, 6(1):6, 2023.

\bibitem{tcgalusc}
S Kirk, Y Lee, P Kumar, et~al.
\newblock The cancer genome atlas lung squamous cell carcinoma collection (tcga-lusc), version 4 [dataset].
\newblock {\em The Cancer Imaging Archive}, 2016.

\bibitem{KOENIG2020102248}
Lauren~N. Koenig, Gregory~S. Day, Amber Salter, Sarah Keefe, Laura~M. Marple, Justin Long, Pamela LaMontagne, Parinaz Massoumzadeh, B.~Joy Snider, Manasa Kanthamneni, Cyrus~A. Raji, Nupur Ghoshal, Brian~A. Gordon, Michelle Miller-Thomas, John~C. Morris, Joshua~S. Shimony, and Tammie~L.S. Benzinger.
\newblock Select atrophied regions in alzheimer disease (sara): An improved volumetric model for identifying alzheimer disease dementia.
\newblock {\em NeuroImage: Clinical}, 26:102248, 2020.

\bibitem{kollias2021mia}
Dimitrios Kollias, Anastasios Arsenos, Levon Soukissian, and Stefanos Kollias.
\newblock Mia-cov19d: Covid-19 detection through 3-d chest ct image analysis.
\newblock In {\em Proceedings of the IEEE/CVF International Conference on Computer Vision}, pages 537--544, 2021.

\bibitem{korolev2017residual}
S Korolev, A Safiullin, M Belyaev, and Y Dodonova.
\newblock {Residual and plain convolutional neural networks for 3D brain MRI classification}.
\newblock In {\em Proceedings-International Symposium on Biomedical Imaging}, pages 835--838, 2017.

\bibitem{LaMontagne2019}
Pamela~J. LaMontagne, Tammie~LS. Benzinger, John~C. Morris, Sarah Keefe, Russ Hornbeck, Chengjie Xiong, Elizabeth Grant, Jason Hassenstab, Krista Moulder, Andrei~G. Vlassenko, Marcus~E. Raichle, Carlos Cruchaga, and Daniel Marcus.
\newblock Oasis-3: Longitudinal neuroimaging, clinical, and cognitive dataset for normal aging and alzheimer disease.
\newblock {\em medRxiv}, 2019.

\bibitem{lee2017curated}
Rebecca~Sawyer Lee, Francisco Gimenez, Assaf Hoogi, Kanae~Kawai Miyake, Mia Gorovoy, and Daniel~L Rubin.
\newblock A curated mammography data set for use in computer-aided detection and diagnosis research.
\newblock {\em Scientific Data}, 4(1):1--9, 2017.

\bibitem{lin2023pmc}
Weixiong Lin, Ziheng Zhao, Xiaoman Zhang, Chaoyi Wu, Ya Zhang, Yanfeng Wang, and Weidi Xie.
\newblock Pmc-clip: Contrastive language-image pre-training using biomedical documents.
\newblock In {\em Medical Image Computing and Computer Assisted Intervention -- MICCAI}, 2023.

\bibitem{tcgabrca}
W Lingle et~al.
\newblock The cancer genome atlas breast invasive carcinoma collection (tcga-brca) (version 3) [data set].
\newblock {\em The Cancer Imaging Archive}, 2016.

\bibitem{liu2021automatic}
Chengeng Liu and Qingshan Yin.
\newblock Automatic diagnosis of covid-19 using a tailored transformer-like network.
\newblock In {\em Journal of Physics: Conference Series}, number~1, page 012175. IOP Publishing, 2021.

\bibitem{liu2020chestxdet10}
Jingyu Liu, Jie Lian, and Yizhou Yu.
\newblock Chestx-det10: Chest x-ray dataset on detection of thoracic abnormalities, 2020.

\bibitem{majkowska2020chest}
Anna Majkowska et~al.
\newblock Chest radiograph interpretation with deep learning models: assessment with radiologist-adjudicated reference standards and population-adjusted evaluation.
\newblock {\em Radiology}, 294(2):421--431, 2020.

\bibitem{Marcus2007}
Daniel~S. Marcus, Tracy~H. Wang, Jamie Parker, John~G. Csernansky, John~C. Morris, and Randy~L. Buckner.
\newblock {Open Access Series of Imaging Studies (OASIS): Cross-sectional MRI Data in Young, Middle Aged, Nondemented, and Demented Older Adults}.
\newblock {\em Journal of Cognitive Neuroscience}, 19(9):1498--1507, 09 2007.

\bibitem{mei2022radimagenet}
Xueyan Mei, Zelong Liu, Philip~M Robson, Brett Marinelli, Mingqian Huang, Amish Doshi, Adam Jacobi, Chendi Cao, Katherine~E Link, Thomas Yang, et~al.
\newblock Radimagenet: an open radiologic deep learning research dataset for effective transfer learning.
\newblock {\em Radiology: Artificial Intelligence}, 4(5):e210315, 2022.

\bibitem{morozov2020mosmeddata}
Sergey~P Morozov, AE Andreychenko, NA Pavlov, AV Vladzymyrskyy, NV Ledikhova, VA Gombolevskiy, Ivan~A Blokhin, PB Gelezhe, AV Gonchar, and V~Yu Chernina.
\newblock Mosmeddata: Chest ct scans with covid-19 related findings dataset.
\newblock {\em arXiv preprint arXiv:2005.06465}, 2020.

\bibitem{nguyen2022vindr}
Ha~Q Nguyen, Khanh Lam, Linh~T Le, Hieu~H Pham, Dat~Q Tran, Dung~B Nguyen, et~al.
\newblock {VinDr-CXR: An open dataset of chest X-rays with radiologist’s annotations}.
\newblock {\em Scientific Data}, 9(1):429, 2022.

\bibitem{vindrmammo}
Hieu~T Nguyen, Ha~Q Nguyen, Hieu~H Pham, Khanh Lam, Linh~T Le, Minh Dao, and Van Vu.
\newblock {VinDr-Mammo: A large-scale benchmark dataset for computer-aided diagnosis in full-field digital mammography}.
\newblock {\em Scientific Data}, 10(1):277, 2023.

\bibitem{Nguyen2022VinDrMammoAL}
Hieu~T. Nguyen, Ha~Q. Nguyen, H.~H. Pham, Khanh Lam, L.~T. Le, M. Dao, and Van~H. Vu.
\newblock {VinDr-Mammo: A large-scale benchmark dataset for computer-aided diagnosis in full-field digital mammography}.
\newblock {\em PhysioNet}, 2022.

\bibitem{Nguyen2021VinDrSpineXRAD}
Hieu~T Nguyen, Hieu~H Pham, Nghia~T Nguyen, Ha~Q Nguyen, Thang~Q Huynh, Minh Dao, and Van Vu.
\newblock {VinDr-SpineXR: A deep learning framework for spinal lesions detection and classification from radiographs}.
\newblock In {\em Medical Image Computing and Computer Assisted Intervention--MICCAI 2021: 24th International Conference, Strasbourg, France, September 27--October 1, 2021, Proceedings, Part V 24}, pages 291--301. Springer, 2021.

\bibitem{Nguyen2022VinDrPCXRAO}
Ngoc~Hung Nguyen, Hieu Pham, Thanh-Truong Tran, Tuan Ngoc~Minh Nguyen, and Ha~Q. Nguyen.
\newblock {VinDr-PCXR: An open, large-scale chest radiograph dataset for interpretation of common thoracic diseases in children}.
\newblock {\em PhysioNet}, 2022.

\bibitem{OpenAI2023GPT4TR}
OpenAI.
\newblock {GPT-4 Technical Report}.
\newblock {\em arXiv preprint arXiv:2303.08774}, 2023.

\bibitem{park2021vision}
Sangjoon Park, Gwanghyun Kim, Yujin Oh, Joon~Beom Seo, Sang~Min Lee, Jin~Hwan Kim, Sungjun Moon, Jae-Kwang Lim, and Jong~Chul Ye.
\newblock Vision transformer for covid-19 cxr diagnosis using chest x-ray feature corpus.
\newblock {\em arXiv preprint arXiv:2103.07055}, 2021.

\bibitem{pavlopoulos2019survey}
John Pavlopoulos, Vasiliki Kougia, and Ion Androutsopoulos.
\newblock A survey on biomedical image captioning.
\newblock In {\em Proceedings of the second workshop on shortcomings in vision and language}, pages 26--36, 2019.

\bibitem{pavlova2021covid}
Maya Pavlova et~al.
\newblock Covid-net cxr-2: An enhanced deep convolutional neural network design for detection of covid-19 cases from chest x-ray images.
\newblock {\em Frontiers in Medicine}, 9, 2022.

\bibitem{Pedrosa2019LNDbAL}
Jo{\~a}o Pedrosa, Guilherme Aresta, Carlos~A. Ferreira, M{\'a}rcio Rodrigues, Patr{\'i}cia Leit{\~a}o, Andr{\'e} Carvalho, Jo{\~a}o Rebelo, Eduardo Negr{\~a}o, Isabel Ramos, Ant{\'o}nio Cunha, and Aur{\'e}lio J.~C. Campilho.
\newblock Lndb: A lung nodule database on computed tomography.
\newblock {\em ArXiv}, abs/1911.08434, 2019.

\bibitem{Qiu_2017_ICCV}
Zhaofan Qiu, Ting Yao, and Tao Mei.
\newblock Learning spatio-temporal representation with pseudo-3d residual networks.
\newblock In {\em Proceedings of the IEEE International Conference on Computer Vision (ICCV)}, Oct 2017.

\bibitem{rahman2021exploring}
Tawsifur Rahman, Amith Khandakar, Yazan Qiblawey, Anas Tahir, Serkan Kiranyaz, Saad Bin~Abul Kashem, Mohammad~Tariqul Islam, Somaya Al~Maadeed, Susu~M Zughaier, Muhammad~Salman Khan, et~al.
\newblock Exploring the effect of image enhancement techniques on covid-19 detection using chest x-ray images.
\newblock {\em Computers in biology and medicine}, 132:104319, 2021.

\bibitem{rajpurkar2017mura}
Pranav Rajpurkar, Jeremy Irvin, Aarti Bagul, Daisy Ding, Tony Duan, Hershel Mehta, Brandon Yang, Kaylie Zhu, Dillon Laird, Robyn~L Ball, et~al.
\newblock Mura: Large dataset for abnormality detection in musculoskeletal radiographs.
\newblock {\em arXiv preprint arXiv:1712.06957}, 2017.

\bibitem{DVN/6ACUZJ_2021}
Shokouh Shakouri, Mohammad~Amin Bakhshali, Parvaneh Layegh, Behzad Kiani, Farid Masoumi, Saeedeh Ataei~Nakhaei, and Sayyed~Mostafa Mostafavi.
\newblock Covid19-ct-dataset: an open-access chest ct image repository of 1000+ patients with confirmed covid-19 diagnosis.
\newblock {\em BMC Research Notes}, 14(1):178, 2021.

\bibitem{shih2019augmenting}
George Shih, Carol~C Wu, Safwan~S Halabi, Marc~D Kohli, Luciano~M Prevedello, et~al.
\newblock Augmenting the national institutes of health chest radiograph dataset with expert annotations of possible pneumonia.
\newblock {\em Radiology: Artificial Intelligence}, 1(1):e180041, 2019.

\bibitem{vstajduhar2017semi}
Ivan {\v{S}}tajduhar, Mihaela Mamula, Damir Mileti{\'c}, and Goezde Uenal.
\newblock Semi-automated detection of anterior cruciate ligament injury from mri.
\newblock {\em Computer methods and programs in biomedicine}, 140:151--164, 2017.

\bibitem{swati2019brain}
Zar Nawab~Khan Swati, Qinghua Zhao, Muhammad Kabir, Farman Ali, Zakir Ali, Saeed Ahmed, and Jianfeng Lu.
\newblock Brain tumor classification for mr images using transfer learning and fine-tuning.
\newblock {\em Computerized Medical Imaging and Graphics}, 75:34--46, 2019.

\bibitem{tiu2022expert}
Ekin Tiu, Ellie Talius, Pujan Patel, Curtis~P Langlotz, Andrew~Y Ng, and Pranav Rajpurkar.
\newblock Expert-level detection of pathologies from unannotated chest x-ray images via self-supervised learning.
\newblock {\em Nature Biomedical Engineering}, 6(12):1399--1406, 2022.

\bibitem{tu2023towards}
Tao Tu, Shekoofeh Azizi, Danny Driess, Mike Schaekermann, Mohamed Amin, Pi-Chuan Chang, Andrew Carroll, Chuck Lau, Ryutaro Tanno, Ira Ktena, et~al.
\newblock Towards generalist biomedical ai.
\newblock {\em arXiv preprint arXiv:2307.14334}, 2023.

\bibitem{wang2020score}
Haofan Wang, Zifan Wang, Mengnan Du, Fan Yang, Zijian Zhang, Sirui Ding, Piotr Mardziel, and Xia Hu.
\newblock Score-cam: Score-weighted visual explanations for convolutional neural networks.
\newblock In {\em Proceedings of the IEEE/CVF conference on computer vision and pattern recognition workshops}, pages 24--25, 2020.

\bibitem{wang2017chestx}
Xiaosong Wang, Yifan Peng, Le Lu, Zhiyong Lu, Mohammadhadi Bagheri, and Ronald~M Summers.
\newblock Chestx-ray8: Hospital-scale chest x-ray database and benchmarks on weakly-supervised classification and localization of common thorax diseases.
\newblock In {\em Proceedings of the IEEE/CVF Conference on Computer Vision and Pattern Recognition (CVPR)}, pages 2097--2106, 2017.

\bibitem{wang2022self}
Xuezhi Wang, Jason Wei, Dale Schuurmans, Quoc Le, Ed Chi, Sharan Narang, Aakanksha Chowdhery, and Denny Zhou.
\newblock Self-consistency improves chain of thought reasoning in language models.
\newblock {\em arXiv preprint arXiv:2203.11171}, 2022.

\bibitem{Wu2023KDiagKD}
Chaoyi Wu, Xiaoman Zhang, Yanfeng Wang, Ya Zhang, and Weidi Xie.
\newblock K-diag: Knowledge-enhanced disease diagnosis in radiographic imaging.
\newblock {\em Medical Image Computing and Computer Assisted Intervention -- MICCAI Workshop}, 2023.

\bibitem{Wu2023MedKLIPMK}
Chaoyi Wu, Xiaoman Zhang, Ya Zhang, Yanfeng Wang, and Weidi Xie.
\newblock {MedKLIP: Medical Knowledge Enhanced Language-Image Pre-Training}.
\newblock {\em \change{IEEE International Conference on Computer Vision (ICCV)}}, 2023.

\bibitem{Wu2023TowardsGF}
Chaoyi Wu, Xiaoman Zhang, Ya Zhang, Yanfeng Wang, and Weidi Xie.
\newblock Towards generalist foundation model for radiology by leveraging web-scale 2d\&3d medical data.
\newblock {\em \change{arXiv preprint arXiv:2308.02463}}, 2023.

\bibitem{yan2017deeplesion}
Ke Yan, Xiaosong Wang, Le Lu, and Ronald~M Summers.
\newblock Deeplesion: Automated deep mining, categorization and detection of significant radiology image findings using large-scale clinical lesion annotations.
\newblock {\em arXiv preprint arXiv:1710.01766}, 2017.

\bibitem{zhang2020clinically}
Kang Zhang, Xiaohong Liu, Jun Shen, Zhihuan Li, Ye Sang, Xingwang Wu, Yunfei Zha, Wenhua Liang, Chengdi Wang, Ke Wang, et~al.
\newblock Clinically applicable ai system for accurate diagnosis, quantitative measurements, and prognosis of covid-19 pneumonia using computed tomography.
\newblock {\em Cell}, 181(6):1423--1433, 2020.

\bibitem{zhang2023biomedclip}
Sheng Zhang, Yanbo Xu, Naoto Usuyama, Hanwen Xu, Jaspreet Bagga, Robert Tinn, Sam Preston, Rajesh Rao, Mu Wei, Naveen Valluri, et~al.
\newblock Biomedclip: a multimodal biomedical foundation model pretrained from fifteen million scientific image-text pairs.
\newblock {\em arXiv preprint arXiv:2303.00915}, 2023.

\bibitem{zhang2023knowledge}
Xiaoman Zhang, Chaoyi Wu, Ya Zhang, Yanfeng Wang, and Weidi Xie.
\newblock Knowledge-enhanced pre-training for auto-diagnosis of chest radiology images.
\newblock {\em Nature Communications}, 2023.

\bibitem{Zhang2023PMCVQAVI}
Xiaoman Zhang, Chaoyi Wu, Ziheng Zhao, Weixiong Lin, Ya Zhang, Yanfeng Wang, and Weidi Xie.
\newblock Pmc-vqa: Visual instruction tuning for medical visual question answering.
\newblock {\em arXiv preprint arXiv:2305.10415}, 2023.

\end{thebibliography}

\section{ACKNOWLEDGEMENT}
This work is supported by the National Key R\&D Program of China (No. 2022ZD0160702), STCSM (No. 22511106101, No. 18DZ2270700, No. 21DZ1100100), 111 plan (No. BP0719010), and State Key Laboratory of UHD Video and Audio Production and Presentation. Additionally, We sincerely thank all the contributors who uploaded the relevant data to Radiopaedio. We appreciate their willingness to make these valuable cases publicly available, promoting the related research greatly.

\section{AUTHOR CONTRIBUTIONS}
All listed authors clearly meet the ICMJE 4 criteria. Specifically, Qiaoyu Zheng, Weike Zhao, Chaoyi Wu, Xiaoman Zhang, Lisong Dai, Hengyu Guan, Yuehua Li, Ya Zhang, Yanfeng Wang, and Weidi Xie all make contributions to the conception or design of the work, and Qiaoyu Zheng, Weike Zhao, Chaoyi Wu further perform acquisition, analysis, or interpretation of data for the work. Lisong Dai and Yuehua Li as experienced clinicians help check the correctness of cases and confirm the medical related knowledge. In writing, Qiaoyu Zheng, Weike Zhao, Chaoyi Wu draft the work and Xiaoman Zhang, Lisong Dai, Hengyu Guan, Yuehua Li, Ya Zhang, Yanfeng Wang, and Weidi Xie review it critically for important intellectual content. All authors approve of the version to be published and agree to be accountable for all aspects of the work in ensuring that questions related to the accuracy or integrity of any part of the work are appropriately investigated and resolved.  Qiaoyu Zheng, Weike Zhao, Chaoyi Wu contribute equally to the work. Weidi Xie and Yanfeng Wang are the corresponding authors.

\section{COMPETING INTERESTS}
The authors declare no competing interests.

% \clearpage
% \appendix
\newpage
\section{Supplementary}
\label{supp}

\captionsetup[figure]{name=Supplementary Figure}
\captionsetup[table]{name=Supplementary Table}
\setcounter{figure}{0}
\setcounter{table}{0}

\subsection{Implementation Details}
At training time, we consider two diagnosis tasks in different granularities: 
Disorder-level classification~(5569 classes) and ICD10-level classification~(931 Classes). The image input will all be resized to $256\times256 \times D$ in height, width, and depth respectively. For 3D scans, the depth is treated as a factor for ablation study, and will be discussed in the experiment section. In the vision encoder, we adopt two separate modules for normalization and a shared module to compute the final embedding. The detailed architecture design will also be further discussed in our ablation study. In fusion module, we use a 2-layer, 4-head transformer encoder with a learnable ``[CLS]'' token for final prediction. 

For augmentation, we adopt Gaussian Noise, Contrast Adjustment, Affine Variation, 
and Elastic Deformation, implemented from the MONAI~\cite{cardoso2022monai} package. 
For optimization, we utilize the AdamW optimizer with a cosine learning rate curve, setting the maximum learning rate to $lr=1\times e^{-5}$, with an adjustable batch size ranging from 4 to 32 depending on the input image depth and model scale to avoid out-of-memory error. The total training duration spans 100 epochs, with the initial 5 epochs for warm-up. 
At fine-tuning stage, we adopt a similar model architecture and optimization setting. 

\subsection{Fusion Strategy}
\label{Fuse}
In this part, we will discuss the fusion strategy for baseline implementation in detail and \change{this will be adopted for all the compared methods that are without fusion module.}. Since most existing architectures cannot perform case-level diagnosis, we propose three parameter-free methods to give case-level prediction with a scan-level classifier:
\vspace{-0.2cm}
\begin{itemize}
    \setlength\itemsep{0.1cm}
    \item \textbf{Random Picking.} We randomly select an image scan a case. It is based on the belief that each scan in a case should encompass the essential information required for an accurate.
    \item \textbf{Max Pooling.} We adopt max pooling on all the images in a case to get the final prediction. It is based on the principle that if there is an image within a case that can illustrate the presence of abnormality, we should consider it as an unhealthy case.
    \item  \textbf{Mean Pooling.} Instead of max pooling, we replace it with mean pooling in this case. It is rooted in the notion that relying solely on the diagnosis from a single image is insufficient, and our goal is to comprehensively leverage all the scan image information in a case.
\end{itemize}

Subsequently, we will present a comparative analysis of the outcomes obtained through these strategies, focusing on the two classification standards, ICD-10-CM and disorders. As shown in the Supplementary Table~\ref{supple_figure:1} and Supplementary Table~\ref{supple_figure:2}, in most cases ``Max Pooling'' outperform others. Therefore, in the main paper, we adopt max pooling as our baseline for comparison.

\begin{table}[!htb]
\centering
\footnotesize
\vspace{-0.2cm}
\setlength{\tabcolsep}{8pt}
\caption{\textbf{Results on ICD10-CM classes. We report the different scores for Head/Medium/Tail classes, ``Rand'', ``Max''.} ``Mean'' represent the random picking, Max Pooling and Mean Pooling respectively.}
\vspace{3pt}
\begin{tabular}{cccccccccc}
\toprule
\multirow{2}{*}{\textbf{Granularity}}            & \multirow{2}{*}{\textbf{Split}}              & \multirow{2}{*}{\textbf{Mode}} & \multicolumn{7}{c}{\textbf{Metrics}}                                                                                 \\
                                                 &                                              &                                       & AUC            & AP             & F1             & MCC            & R@0.01         & R@0.05         & R@0.1          \\ \midrule
\multicolumn{1}{c|}{\multirow{9}{*}{ICD-10-CM}} & \multicolumn{1}{c|}{\multirow{3}{*}{Head}}   & \multicolumn{1}{c|}{Rand} & 89.12          & 12.39          & 20.51          & 21.72          & 24.12          & 51.82          & 67.93          \\
\multicolumn{1}{c|}{}                            & \multicolumn{1}{c|}{}                        & \multicolumn{1}{c|}{ Max}  & \textbf{90.39}          & 12.80          & \textbf{20.65}          & \textbf{21.94}          & \textbf{24.39}          & \textbf{54.12}          & \textbf{70.11}          \\
\multicolumn{1}{c|}{}                            & \multicolumn{1}{c|}{}                        & \multicolumn{1}{c|}{Mean} & 85.49          & \textbf{13.05}          & 20.25          & 21.52          & 24.31          & 49.92          & 65.28          \\ \cmidrule{2-10} 
\multicolumn{1}{c|}{}                            & \multicolumn{1}{c|}{\multirow{3}{*}{Medium}} & \multicolumn{1}{c|}{Rand} & 88.98          & 7.72           & \textbf{15.44}          & 18.02          & 22.52          & 50.50          & 63.96          \\
\multicolumn{1}{c|}{}                            & \multicolumn{1}{c|}{}                        & \multicolumn{1}{c|}{Max}  & \textbf{90.15} & \textbf{7.74}           & 15.41          & \textbf{17.96}          & \textbf{22.75}          & \textbf{51.10}          & \textbf{66.31}          \\
\multicolumn{1}{c|}{}                            & \multicolumn{1}{c|}{}                        & \multicolumn{1}{c|}{Mean} & 85.05          & 7.69           & 15.03          & 17.53          & 22.58          & 47.41          & 64.52          \\ \cmidrule{2-10} 
\multicolumn{1}{c|}{}                            & \multicolumn{1}{c|}{\multirow{3}{*}{Tail}}   & \multicolumn{1}{c|}{Rand} & 86.41          & \textbf{4.46}           & \textbf{8.54}  & \textbf{12.48} & 8.37           & 23.74          & 37.12          \\
\multicolumn{1}{c|}{}                            & \multicolumn{1}{c|}{}                        & \multicolumn{1}{c|}{Max}  & \textbf{86.46} & 2.90           & 6.09           & 10.27          & \textbf{8.51}           & \textbf{25.06} & \textbf{42.80} \\
\multicolumn{1}{c|}{}                            & \multicolumn{1}{c|}{}                        & \multicolumn{1}{c|}{Mean} & 81.50          & 2.68           & 5.42           & 9.04           & 6.15           & 20.13          & 30.64          \\ \bottomrule
\end{tabular}
\label{supple_figure:1}
\end{table}

\begin{table}[!htb]
\centering
\footnotesize
\setlength{\tabcolsep}{8pt}
\caption{\textbf{Results on Disorder classes.} We report the different scores for Head/Medium/Tail classes, ``Rand'', ``Max''. ``Mean'' represent the random picking, Max Pooling and Mean Pooling respectively.}
\vspace{3pt}
\begin{tabular}{cccccccccc}
\toprule
\multirow{2}{*}{\textbf{Granularity}}            & \multirow{2}{*}{\textbf{Split}} & \multirow{2}{*}{\textbf{Mode}} & \multicolumn{7}{c}{\textbf{Metrics}}                                                                                                    \\
                                                 &                                 &                                       & AUC            & AP                                & F1             & MCC            & R@0.01         & R@0.05         & R@0.1          \\ \midrule
\multicolumn{1}{c|}{\multirow{9}{*}{Disorders}} & \multicolumn{1}{c|}{\multirow{3}{*}{Head}}          & \multicolumn{1}{c|}{Rand} & 91.32          & 14.08                             & 22.93          & \textbf{24.61}          & 30.68          & 58.78          & 73.79          \\
\multicolumn{1}{c|}{}                            &          \multicolumn{1}{c|}{}                          & \multicolumn{1}{c|}{Max}  & \textbf{91.91}          & 13.56                             & 22.08          & 23.66          & 30.88          & \textbf{60.50}          & \textbf{74.70}          \\
\multicolumn{1}{c|}{}                            &           \multicolumn{1}{c|}{}                         & \multicolumn{1}{c|}{Mean} & 88.72          & \textbf{14.29}                    & \textbf{23.26}          & 24.60          & \textbf{31.97}          & 59.67          & 71.22          \\\cmidrule{2-10} 
\multicolumn{1}{c|}{}                            & \multicolumn{1}{c|}{\multirow{3}{*}{Medium}}         & \multicolumn{1}{c|}{Rand} & 92.63          & 9.47                              & \textbf{17.62}          & \textbf{20.61}          & 28.18          & 57.04          & 71.42          \\
\multicolumn{1}{c|}{}                            &            \multicolumn{1}{c|}{}                        & \multicolumn{1}{c|}{Max}  & \textbf{93.62}          & 9.19                              & 17.16          & 20.42          & 28.24          & \textbf{59.32}          & \textbf{73.80}          \\
\multicolumn{1}{c|}{}                            &       \multicolumn{1}{c|}{}                             & \multicolumn{1}{c|}{Mean} & 90.97          & \textbf{9.54}                              & 17.33          & 20.42          & \textbf{29.31}          & 58.15          & 71.00          \\ \cmidrule{2-10} 
\multicolumn{1}{c|}{}                            & \multicolumn{1}{c|}{\multirow{3}{*}{Tail}}          & \multicolumn{1}{c|}{Rand} & 87.61          & \textbf{4.85}                              & \textbf{8.66}           & \textbf{13.26}          & 8.58           & 24.25          & 41.39          \\
\multicolumn{1}{c|}{}                            &         \multicolumn{1}{c|}{}                           & \multicolumn{1}{c|}{Max}  & \textbf{88.55}          & 4.34                              & 8.03           & 12.96          & 9.13           & \textbf{27.46} & 41.43          \\
\multicolumn{1}{c|}{}                            &         \multicolumn{1}{c|}{}                           & \multicolumn{1}{c|}{Mean} & 84.50          & 4.35                              & 8.09           & 12.80          & \textbf{9.38}           & 25.96          & \textbf{42.57} \\\bottomrule
\end{tabular}
\label{supple_figure:2}
\end{table}

\subsection{Ablation Study}
\label{ablation}
To explore the optimal model architecture and parameter configurations, including the fusion strategy, visual
encoder architecture, the depth of 3D scans, and augmentation. The results on the visual
encoder architecture can be found in the main body.
In the default experiment setting, we set 512x512 as the image size, use 16 as the 3D scan depth, and ResNet-18+ResNet-34 as the visual backbone, without knowledge enhancement and augmentation strategy. While conducting an ablation study on certain component, we keep other settings unchanged.

\noindent \textbf{Image Dimension.}
%\qiaoyu{Settings: no augmentation, BCE}
Here, we aim to investigate the effect of input resolution, {\em i.e.}, increasing the depth of input volumes. 
Due to the constraints of GPU physical memory, we experiment with 16, 24, and 32 as the depths for 3D input volume. We employ trilinear interpolation to resample 3D scans to the same size. 
Supplementary Table~\ref{tab:Image Depth Comparison} illustrates the performance by varying the depths of 3D images. It can be observed that an increase in depth can bring clear performance gain, suggesting that detailed depth information is critical to perform diagnosis. Consequently, with more slices, the model tends to yield more favorable results.

\begin{table}[!htb]
  \centering
  \small
  \setlength{\tabcolsep}{12pt}
  \caption{\textbf{Ablation study on different 3D scan depth size settings.} Considering the computational cost, we evaluated the commonly seen 16, 24, 32 depth sizes on our datasets.}
  \vspace{3pt}
\begin{tabular}{c|ccccccc}
\toprule
3D Image Depth & AUC   & AP    & F1    & MCC   & R@0.01 & R@0.05 & R@0.1 \\ \midrule
16             & 87.06 & 11.27 & 17.36 & 19.23 & 21.48  & 44.38  & 61.54 \\
24             & 87.49 & 11.41 & 17.84 & 19.39 & 21.76  & 45.72  & 62.53 \\
\textbf{32} & \textbf{88.13} & \textbf{11.98} & \textbf{18.02} & \textbf{19.67} & \textbf{22.84} & \textbf{47.46} & \textbf{63.91} \\ \bottomrule
\end{tabular}
\label{tab:Image Depth Comparison}
\end{table}

\noindent \textbf{Augmentation.}
%\qiaoyu{BCE, depth=16}
%In the literaAugmentation strategy has been proved to be a simple but mostly effective strategy when facing long-tailed or few-shot classification tasks. 
Here, we evaluate the effectiveness of augmentation on our dataset. Specifically, we adopt four augmentation strategies with $15\%$ probability each, namely, Gaussian Noise, Contrast Adjustment, Affine Variation, and Elastic Deformation. As a result, about half of the training data will be applied at least one augmentation in each training batch. As shown in Supplementary Table~\ref{Tab:Augmentation Comparison}, adopting data augmentation shows notable performance improvement. This improvement is particularly evident on AUC and AP when employing the ViT as the visual backbone while, still, ResNet-based model performs better.

\begin{table}[!htb]
\centering
\small
\setlength{\tabcolsep}{8pt}
\caption{\textbf{Ablation on adopting augmentation strategy.} In the table, we show the effectiveness of adopting reasonable data augmentation for our task. ``ResNet'' denotes the ResNet-based visual encoder architecture and ``ViT'' denotes the ViT-basde one.}
\vspace{3pt}
\begin{tabular}{cc|ccccccc}
\toprule
Augmentation  & Visual Encoder & AUC   & AP    & F1    & MCC   & R@F0.01 & R@F0.05 & R@F0.1 \\ \midrule
\multirow{2}{*}{\textbf{YES}} & ResNet         & 88.96 & 13.92 & 22.04 & 23.08 & 25.75   & 51.61   & 66.47  \\
                              & ViT            & 84.22 & 8.69  & 16.36 & 17.18 & 19.83   & 38.70   & 54.04  \\ \midrule
\multirow{2}{*}{NO}           & ResNet         & 87.06 & 11.27 & 17.36 & 19.23 & 21.48   & 44.38   & 61.54  \\
                              & ViT            & 81.69 & 6.40  & 14.71 & 15.30 & 18.11   & 34.73   & 48.84  \\ \bottomrule
\end{tabular}
\label{Tab:Augmentation Comparison}
\end{table}

In conclusion, we find the ResNet-based model with augmentation strategy and unifying the 3D scan depth to 32 are the most suitable settings for our long-tailed case-level multi-modal diagnosis task.

\subsection{\change{In-class Analysis}}
\change{
The experimental results on whether to use fusion strategy and knowledge enhancement strategy show that, statistically, models that combine both strategies tend to have a higher average AUC than those that do not, across the head, medium, and tail categories, with some categories showing noticeable increase, while some may remain same or slightly decreasing. Taking certain categories within the head category as examples from Supplementary Figure~\ref{fig:In-class visulization}. The application of the fusion strategy do not result in improvements for the Meningioma class. Similarly, the implementation of the Knowledge Enhancement strategy do not lead to improvements for the Renal cell carcinoma clas. However, we can observe significant improvements in more classes such as Normal, Tuberculosis, and Lymphoma. Consequently, the average metrics across classes shows a notable enhancement.}

%in AUC
%However, in practise, employing these two strategies does not lead to improvements in all categories' AUC metrics uniformly. 
%Some categories show a noticeable increase in AUC, while others may experience a decrease. \weidi{this is too negative, reword it. }
%It's just that the degree of improvement statistically outweighs the degree of decrease, which ultimately manifests as a significant gain in the average results.}
\subsection{\change{Comparison with Current Diagnosis Datasets}}
\label{Supple:Data_comparison}
\change{In this section, we provide more detailed comparison with other current public available diagnosis datasets as shown in Supplementary Table~\ref{tab:dataset}. From the table, we can see our proposed dataset RP3D-DiagDS is the first diagnosis dataset considering whole body regions across different imaging modalities and cover a wide range of diseases.}

\begin{table}[!t]
\centering
\small
\setlength{\tabcolsep}{2pt}
\renewcommand{\arraystretch}{1.0}
\begin{threeparttable}
\caption{\textbf{Comparison between RP3D-DiagDS and existing disease diagnosis datasets.}
\#Image refers to the number of images in the dataset.
\#Disorders represents the number of categories of labeled disorders.
2D/3D indicates the dimension of images in the dataset, and the case-level column indicates whether the dataset contains multiple images per patient.
Our dataset demonstrates a significant increase in both size and diversity.}
\label{tab:dataset}
\vspace{3pt}
\begin{tabular}{lccccccc}
\toprule
Dataset   & Anatomy & Modality & \#Image &  \#Disorders & 2D & 3D  &  Case-level \\
\midrule

% LDCT~\cite{moen2021low} & Head, Chest, Abdomen & CT & 25k & \weike{-} &  \xmark & \checkmark &\xmark\\
\change{ADNI}~\cite{jack2008alzheimer} & Head and Neck & MRI & 112k & 4  &   \xmark & \checkmark&\xmark\\
Brain-Tumor~\cite{msoud_nickparvar_2021} & Head and Neck & MRI &  7k & 4 &   \checkmark & \xmark&\xmark\\

% Brain-MRI-Scans ~\cite{SHREYA_2023} & Head & MRI &  1.3k & 4 &    \checkmark & \xmark &\xmark\\
CE-MRI ~\cite{cheng2015enhanced} & Head and Neck & MRI &  3k & 3 &    \checkmark & \xmark &\xmark\\

Brain-Tumor-17 ~\cite{FERNANDO_2023} & Head and Neck & MRI &  4.4k & 17 &    \checkmark & \xmark &\xmark\\
OASIS\cite{KOENIG2020102248,LaMontagne2019,Marcus2007} & Head and Neck & MRI &  3.2k & 2 &    \checkmark & \xmark &\xmark\\
ICH2020~\cite{data5010014} & Head and Neck & CT &  82 & 5 &  \xmark & \checkmark  &\xmark\\
COVID19-CT-DB~\cite{DVN/6ACUZJ_2021} & Chest & CT & 1k & 9 &    \checkmark & \xmark &\xmark\\
\change{MosMedData}~\cite{morozov2020mosmeddata} & Chest & CT & 1.1k & 5  &   \xmark & \checkmark&\xmark\\
\change{LNDb}~\cite{Pedrosa2019LNDbAL} & Chest & CT & 237 & 3  &   \xmark & \checkmark&\xmark\\
\change{NSCLC}~\cite{bakr2018radiogenomic} & Chest & CT, Nuclear Medicine & 211 & 4  &   \xmark & \checkmark&\xmark\\
Covid-CXR2~\cite{pavlova2021covid} & Chest & X-ray & 19k & 3  &   \checkmark & \xmark &\xmark\\
POCUS~\cite{born2020pocovid} & Chest & Ultrasound & 1.1k & 3  &   \checkmark & \xmark &\xmark\\
MIA-COV19~\cite{kollias2021mia}& Chest & CT & 5.3k & 2 &  \xmark & \checkmark  &\xmark\\
\change{CC-CCII}~\cite{zhang2020clinically} & Chest & CT & 3.8k & 3  &  \xmark & \checkmark&\xmark\\
\change{RadChest}~\cite{draelos2021machine} & Chest & CT & 3.6k & 83  &   \checkmark & \xmark&\xmark\\
\change{LIDC-IDRI}~\cite{armato2011lung}& Chest & CT & 1.0k & 3 &   \xmark & \checkmark &\xmark\\

NIH Chestxray~\cite{wang2017chestx} & Chest & X-ray & 100k & 14 &   \checkmark & \xmark &\xmark\\
PadChest~\cite{bustos2020padchest} & Chest & X-ray & 160k & 193 &   \checkmark & \xmark &\xmark\\
CheXpert~\cite{irvin2019chexpert} & Chest & X-ray & 224k & 14  &   \checkmark & \xmark &\xmark\\
% GoogleNIH~\cite{majkowska2020chest} & Chest & X-ray & &   &  2D &\xmark\\

% Object-CXR~\cite{objectcxr} & Chest & X-ray &  10k & 2  &  2D &\xmark\\
% NLM-TB~\cite{jaeger2014two} & Chest & X-ray & 138  &  2 &  2D &\xmark\\
RSNA~\cite{shih2019augmenting} & Chest & X-ray & 30k & 2  &   \checkmark & \xmark &\xmark\\
SIIM-ACR~\cite{filice2020crowdsourcing} & Chest & X-ray & 15k  & 2  &   \checkmark & \xmark &\xmark\\
% OpenI~\cite{demner2016preparing} & Chest & X-ray & 7k &   &  2D &\xmark\\
MIMIC-CXR~\cite{johnson2019mimic} & Chest & X-ray &  371k & 14  &   \checkmark & \xmark &\xmark\\
VinDr-CXR~\cite{nguyen2022vindr} & Chest & X-ray & 18k & 28  &   \checkmark & \xmark &\xmark\\
VinDr-PCXR~\cite{Nguyen2022VinDrPCXRAO} & Chest & X-ray & 9k & 51  &   \checkmark & \xmark&\xmark\\
\change{ChestX-Det10}~\cite{liu2020chestxdet10} & Chest & X-ray & 3.5k & 10  &   \checkmark & \xmark&\xmark\\
\change{COVID-19-Radio}~\cite{chowdhury2020can,rahman2021exploring} & Chest & X-ray & 21k & 4  &   \checkmark & \xmark&\xmark\\
\change{IU-Xray}~\cite{pavlopoulos2019survey} & Chest & X-ray & 7.4k & 2  &   \checkmark & \xmark&\xmark\\
\change{ISPY1}~\cite{hylton2016neoadjuvant} & Breast & MRI & 755 & 5  &   \xmark & \checkmark&\xmark\\
VinDr-Mammo~\cite{Nguyen2022VinDrMammoAL} & Breast & Mammography & 20k &  10 &    \checkmark & \xmark&\xmark\\
DDSM~\cite{DDSM, lee2017curated} & Breast & Mammography &  55k & 5 &    \checkmark & \xmark&\xmark\\
CMMD2022~\cite{cai2019breast} & Breast & Mammography &  5.2k & 2 &    \checkmark & \xmark &\xmark\\
BUSI~\cite{al2020dataset} & Breast & Ultrasound &  780 & 3 &    \checkmark & \xmark &\xmark\\
\change{CT-Kidney}~\cite{islam2022vision} & Abdomen and Pelvis & CT & 12k & 4  &   \checkmark & \xmark&\xmark\\
VinDr-SpineXr~\cite{Nguyen2021VinDrSpineXRAD} & Spine & X-ray & 10k & 13  &    \checkmark & \xmark&\xmark\\
\change{Vertebrase-Xray}~\cite{fraiwan2022using} & Spine & X-ray & 388 & 3 &    \checkmark & \xmark&\xmark\\
MRNet~\cite{bien2018deep}& Limb & MRI & 1370 & 3  &    \checkmark & \xmark&\xmark\\
\change{MURA}~\cite{rajpurkar2017mura} & Limb & X-ray & 15k & 2  &   \checkmark & \xmark&\xmark\\
\change{KneeMRI}~\cite{vstajduhar2017semi} & Limb & X-ray & 917 & 3  &  \checkmark & \xmark&\xmark\\
DeepLesion~\cite{yan2017deeplesion} & Whole body & Radiology & 32k & 22  &    \checkmark & \xmark &\xmark\\
RadImageNet~\cite{mei2022radimagenet} & Whole body & CT, MRI, Ultrasound & 1.35M & 165  &    \checkmark & \xmark &\xmark\\

\midrule
RP3D-DiagDS~(ours) & Whole body & Radiology & 192k & \begin{tabular}[c]{@{}c@{}}5568\\ (ICD-10-CM:930)\tnote{*}\end{tabular}  & \checkmark & \checkmark  & \checkmark \\
\bottomrule
\end{tabular}
\begin{tablenotes}
\item[*] We map disorders to ICD-10-CM for standardization.
\end{tablenotes}
\end{threeparttable}
\end{table}

\subsection{\change{More ROC Curves}}
\label{Supple:ROC}
\change{In this section, we show more ROC curves for the classes in the head class category as shown in Supplementary Figure~\ref{fig:In-class visulization}}

\begin{figure}[htb!]
\includegraphics[width=\textwidth]{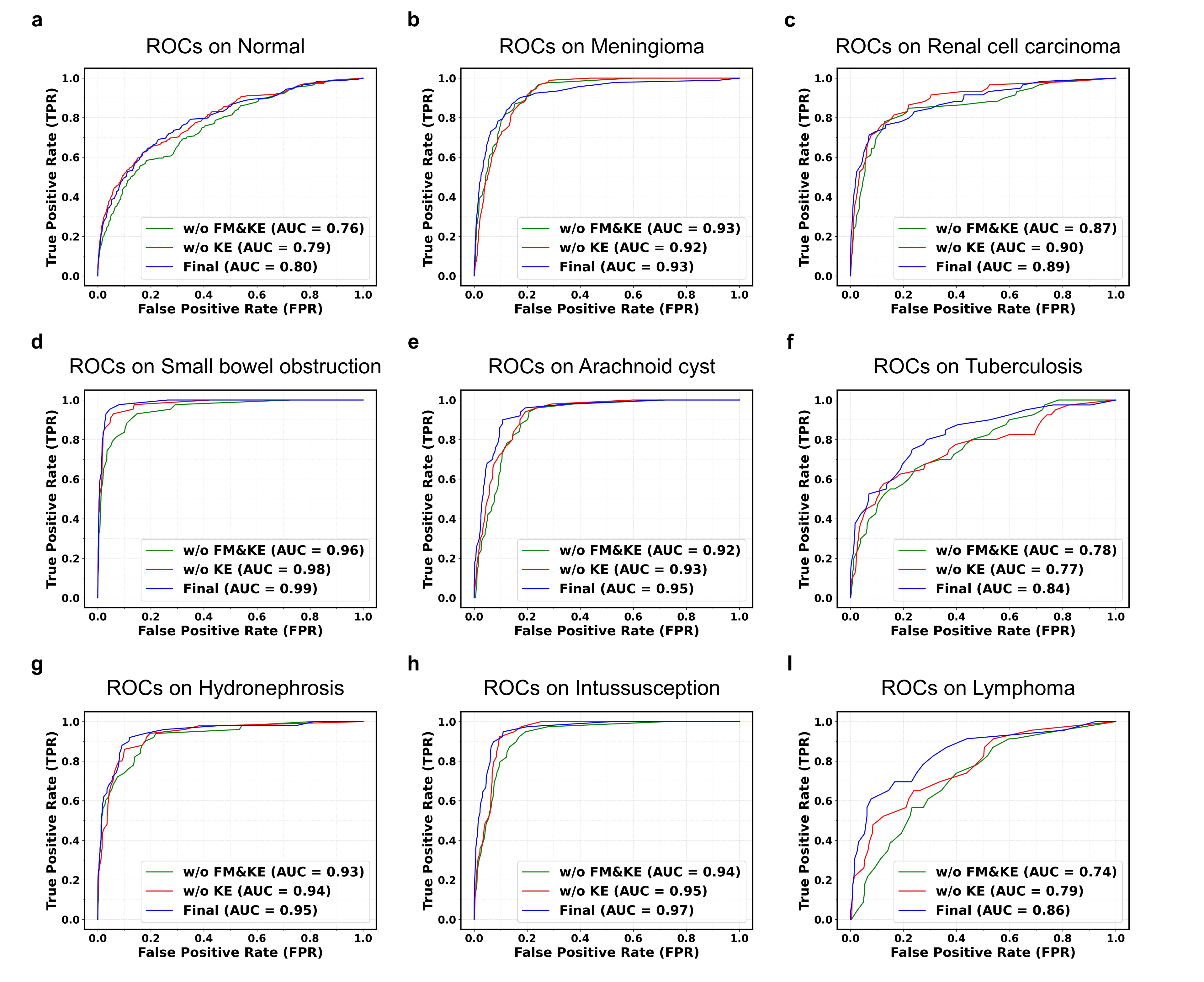}
\caption{\change{\textbf{ROC curves on nine classes in "head" category of disorders.} FM, KE is short for Fusion Module and Knowledge Enhancement.}}
\label{fig:In-class visulization}
\end{figure}

%\end{refsection}
\end{document}